\documentclass[lettersize,journal]{IEEEtran}
\usepackage{times}
\usepackage{epsfig}
\usepackage{graphicx}
\usepackage{amsmath}
\usepackage{amssymb}
\usepackage{times}
\usepackage{graphicx}
\usepackage{verbatim}
\usepackage{multirow}
\usepackage{epsfig}
\usepackage{graphicx}
\usepackage{amsmath}
\usepackage{amssymb}
\usepackage{subfigure}
\usepackage{mathrsfs}
\usepackage{url}
\usepackage{booktabs}
\usepackage{color}
\usepackage{float}
\usepackage{caption}
\usepackage{pifont}
\usepackage{mathrsfs}
\usepackage{bm}
\usepackage{epstopdf}
\usepackage[accsupp]{axessibility} 
\usepackage{titlesec}
\usepackage[switch]{lineno}
\usepackage[pagebackref=true,breaklinks=true,letterpaper=true,colorlinks,citecolor=citecolor,bookmarks=false]{hyperref}
\definecolor{citecolor}{RGB}{34,139,34}

\newcommand{\yx}[1]{\textcolor{black}{\normalsize #1}}
\newcommand{\ycnote}[1]{\textcolor{red}{\small [#1 --Yunchao]}}
\newcommand{\mycaption}[1]{\caption*{\footnotesize #1}}

\newcommand{\ie}{\emph{i.e.}}
\newcommand{\etal}{\textit{et al.}}

\ifCLASSOPTIONcompsoc
  \usepackage[nocompress]{cite}
\else
  \usepackage{cite}
\fi

\ifCLASSINFOpdf
\else
\fi
\begin{document}

%
% paper title
% Titles are generally capitalized except for words such as a, an, and, as,
% at, but, by, for, in, nor, of, on, or, the, to and up, which are usually
% not capitalized unless they are the first or last word of the title.
% Linebreaks \\ can be used within to get better formatting as desired.
% Do not put math or special symbols in the title.
\title{AINet+: Advancing Superpixel Segmentation via Cascaded Association Implantation}
%
%
% author names and IEEE memberships
% note positions of commas and nonbreaking spaces ( ~ ) LaTeX will not break
% a structure at a ~ so this keeps an author's name from being broken across
% two lines.
% use \thanks{} to gain access to the first footnote area
% a separate \thanks must be used for each paragraph as LaTeX2e's \thanks
% was not built to handle multiple paragraphs
%
%
%\IEEEcompsocitemizethanks is a special \thanks that produces the bulleted
% lists the Computer Society journals use for "first footnote" author
% affiliations. Use \IEEEcompsocthanksitem which works much like \item
% for each affiliation group. When not in compsoc mode,
% \IEEEcompsocitemizethanks becomes like \thanks and
% \IEEEcompsocthanksitem becomes a line break with idention. This
% facilitates dual compilation, although admittedly the differences in the
% desired content of \author between the different types of papers makes a
% one-size-fits-all approach a daunting prospect. For instance, compsoc 
% journal papers have the author affiliations above the "Manuscript
% received ..."  text while in non-compsoc journals this is reversed. Sigh.

\author{Yaxiong Wang, Yunchao Wei, Yujiao Wu, Xueming Qian \emph{Member} IEEE, Li Zhu, and Yi Yang% <-this % stops a space
\IEEEcompsocitemizethanks{\
\IEEEcompsocthanksitem Conference version of this work has been accepted by ICCV 2021.
\IEEEcompsocthanksitem Y. Wang  is with the Hefei University of Technology, Hefei, 230000, China Email: wangyx@hfut.edu.cn. \protect
\IEEEcompsocthanksitem Y. Wei  is with the School of Information Science, Beijing Jiaotong University, Beijing, 100000, China. Email: wychao1987@gmail.com. \protect
\IEEEcompsocthanksitem Y. Wu  is with the CSRIO, Australia, Email: yujiaowu111@gmail.com. \protect
\IEEEcompsocthanksitem X. Qian, and L. Zhu is with the Xi'an Jiaotong University, Xi'an, 710049, China. Email: \{qianxm, zhuli\}@mail.xjtu.edu.cn.). \protect
\IEEEcompsocthanksitem Y. Yang  is with the Schoold of Computer Science and Technology, Zhejiang University, Hangzhou, 310000, China, Email:  yangyics@zju.edu.cn. \protect
% <-this % stops an unwanted space
%\IEEEcompsocthanksitem X. Qian is with the Key Laboratory for Intelligent Networks and Network Security, Ministry of Education, Xi’an Jiaotong University, Xi’an 710049, China, also with the SMILES Laboratory, Xi’an Jiaotong University, Xi’an 710049,China, and also with Zhibian Technology Co. Ltd., Taizhou 317000, China.\protect\\
%E-mail:qianxm@mail.xjtu.edu.cn.% <-this % stops an unwanted space
%\IEEEcompsocthanksitem L. Zhu is with the School of Software, Xi’an Jiaotong University, Xi’an 710049, China. \protect\\
%E-mail: zhuli@mail.xjtu.edu.cn.
}
}
\IEEEtitleabstractindextext{%
\begin{abstract}
Superpixel segmentation has seen significant progress benefiting from the deep convolutional networks. The typical approach entails initial division of the image into grids, followed by a learning process that assigns each pixel to adjacent grid segments. However, reliance on convolutions with confined receptive fields results in an implicit, rather than explicit, understanding of pixel-grid interactions. This limitation often leads to a deficit of contextual information during the mapping of associations. To counteract this, we introduce the Association Implantation (AI) module, designed to allow networks to explicitly engage with pixel-grid relationships. This module embeds grid features directly into the vicinity of the central pixel and employs convolutional operations on an enlarged window, facilitating an adaptive transfer of knowledge. This approach enables the network to explicitly extract context at the pixel-grid level, which is more aligned with the objectives of superpixel segmentation than mere pixel-wise interactions. By integrating the AI module across various layers, we enable a progressive refinement of pixel-superpixel relationships from coarse to fine. To further enhance the assignment of boundary pixels, we've engineered a boundary-aware loss function. This function aids in the discrimination of boundary-adjacent pixels at the feature level, thereby empowering subsequent modules to precisely identify boundary pixels and enhance overall boundary accuracy. Our method has been rigorously tested on four benchmarks, including BSDS500, NYUv2, ACDC, and ISIC2017, and our model can achieve competitive performance with comparison methods. 
%Code with detailed guidance is publicly available at \url{https://github.com/wangyxxjtu/AINet-Superpixel}.

\end{abstract}

% Note that keywords are not normally used for peerreview papers.
\begin{IEEEkeywords}
Superpixel segmentation, Segmentation, Deep Learning, Stereo Matching.
\end{IEEEkeywords}}

% make the title area
\maketitle

% To allow for easy dual compilation without having to reenter the
% abstract/keywords data, the \IEEEtitleabstractindextext text will
% not be used in maketitle, but will appear (i.e., to be "transported")
% here as \IEEEdisplaynontitleabstractindextext when the compsoc 
% or transmag modes are not selected <OR> if conference mode is selected 
% - because all conference papers position the abstract like regular
% papers do.
\IEEEdisplaynontitleabstractindextext
% \IEEEdisplaynontitleabstractindextext has no effect when using
% compsoc or transmag under a non-conference mode.

% For peer review papers, you can put extra information on the cover
% page as needed:
% \ifCLASSOPTIONpeerreview
% \begin{center} \bfseries EDICS Category: 3-BBND \end{center}
% \fi
%
% For peerreview papers, this IEEEtran command inserts a page break and
% creates the second title. It will be ignored for other modes.
\IEEEpeerreviewmaketitle

\section{Introduction}
Superpixels are image regions formed by grouping image pixels similar in color and other low-level properties, which could be viewed as an over-segmentation of image. The process of extracting superpixels is known as superpixel segmentation. Comparing to pixels, superpixel provides a more effective representation for image data. With such a compact representation, the computational efficiency of vision algorithms could be improved~\cite{SP_Pool1,supercnn,sp_conv,iccv23,cvpr23,tip241}. Consequently, superpixel could benefit many vision tasks like image recognition~\cite{tip241,tip231}, semantic segmentation~\cite{sp_detect1,sp_detect2,sp_detect3,zheng2020unsupervised,zheng2021rectifying,bmvc23}, object detection~\cite{sp_seg1,sp_seg2}, image denosing~\cite{tmm21}, optical flow estimation~\cite{opt_flow1,opt_flow2,opt_flow3,opt_flow4}, and even adversarial attack~\cite{sp_attack} and domain adaption~\cite{eccv24} can benefit from superpixels. In light of the fundamental importance of superpixels in computer vision, superpixel segmentation attracts much attention since it is first introduced by Ren and Malik~\cite{first_SP} in 2003.

\begin{figure}[t]
\setlength{\abovecaptionskip}{0pt} 
\setlength{\belowcaptionskip}{0pt} 
\begin{center}
    \includegraphics[width=3.3in]{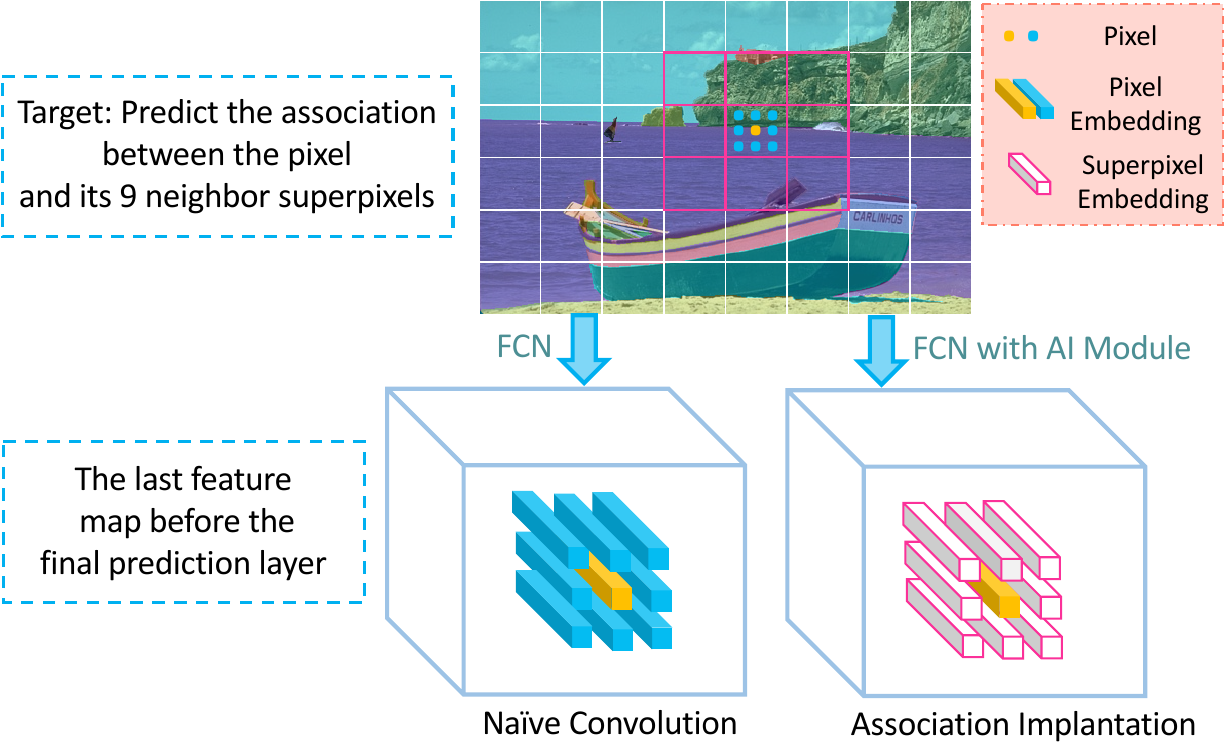}
    %\vspace{0.02cm}
\end{center}
%\vspace{-0.2cm}
\caption{Different from the SCN~\cite{SCN} that implicitly learns the association using the stacked na\"\i ve convolutions, we proposes to implant the corresponding grid features to the surrounding of the pixel to explicitly perceive the relation between each pixel and its neighbor grids.}
\label{first_shown}
\vspace{-0.2cm}
\end{figure}

The common practice for superpixel segmentation is to first split the image into grid cells and then estimate the membership of each pixel to its adjacent grids, by which the grouping could be performed. Meanwhile, the membership estimation plays the key role in superpixel segmentation. Traditional approaches usually utilize the hand-craft features and estimate the relevance of pixel to its neighbor grids based on clustering or graph-based methods~\cite{SLIC,MSLIC,LSC,ERS,SNIC}. For example, the classic method SLIC~\cite{SLIC} treats the surrounding grids as the center candidates and perform clustering procedure on the image. While the ERS~\cite{ERS} models the sueprpixel as a random walk problem, which is also a popular strategy in traditional superpixel segmentation.  However, these methods all suffer from the weakness of the hand-craft features and are difficult to integrate into other trainable deep frameworks.  Inspired by the success of deep neural networks in many computer vision problems, researchers recently attempts to adopt the deep learning technique to superpixel segmentation~\cite{SSN,SCN,tip23,tmm24,tmm19,tmm18}.
As mentioned in abstract, previous deep methods attempt to assign pixels by learning the association of each pixel to its surrounding grids using the fully convolutional networks~\cite{FCN}. The popular solutions like SCN~\cite{SCN}, SSN~\cite{SSN} employ the U-net architecture~\cite{Unet} to predict the association, \ie, the 9-way probabilities, for each pixel. Although stacking convolution layers can enlarge the receptive field and
help study the \emph{pixel-grid wise} probabilities, introducing low-level features with skip connection in the final layer will pollute the probabilities due to the added \emph{pixel-pixel wise} information, since the ultimate target is to predict the association between the target pixel and \emph{its 9-neighbor grids} instead of its 9-neighbor pixels. 

To tackle this weakness, we propose to directly implant the grid features to the surrounding of the corresponding pixel using an association implantation (AI) module. Fig.~\ref{first_shown} simply shows the core idea of our AI module, before feeding the last features into the prediction layer, our AI module is performed: for each pixel, we place the corresponding grid features to its neighbors, then a convolution with $3\times 3$ kernel is followed, this convolution is no longer to capture the pixel-pixel relation but \emph{the relation between pixel and its 9 neighbor grids}, providing the consistent context with the target of superpixel segmentation. Our proposed AI module provides a simple and intuitive way to allow the network to harvest the pixel-neighbor grids context \yx{in an explicit fashion}, which is exactly required by superpixel segmentation. Comparing to existing methods, such a design is more consistent with the target of superpixel segmentation and could give more beneficial support for the subsequent association map inferring. To further boost the performance, we employ our proposed AI module in the penultimate and the last layers, such that a hierarchical pixel-grid relation could be perceived, which could make the performance step further. 

%Thus, the relationships could be learned thanks to the receptive fields, which is an implicit manner to infer the association by iteratively accumulating the context between pixels. However, the cascaded convolutions will progressively incorporate the fine features from the maps of early layers, making the association be falsely predicted as the target pixel and its neighbor pixels rather than the surrounding cells.}

%To address this weakness, we propose to directly implant the grid features to the surrounding of the corresponding pixel using an association implantation (AI) module. Fig.~\ref{first_shown} (b) exhibits the rough pipeline, we first encode
%, whose number of convolution \& pooling operations is adjusted such that the pixels in the output map exactly encodes the features of a grid cell, 
%the hidden features of grid cells. Then, several decoding layers are followed to obtain the pixel-wise embedding. For each pixel embedding, we pad the features of its corresponding neighbor cells to its surrounding, and perform a convolution on the expanded pixel embedding to adaptively propagate the information between the central pixel and its surrounding cells. 
%Our proposed AI module provides a simple and intuitive way to allow the network to harvest the pixel-neighbor cells context \yx{in an explicit fashion}, which is exactly required by superpixel segmentation. Comparing to existing methods, such a design is more consistent with the target of superpixel segmentation and could give more beneficial support for the subsequent association map inferring. 

Besides, a satisfactory superpixel algorithm should accurately identify the boundary pixels, however, some designs towards this target still missed among existing works. To pursue better boundary precision, 
%we propose a boundary-perceiving loss to highlight the image boundaries 
\yx{we augment the optimization with a boundary-perceiving loss}. To be specific, we first sample a set of small local patches on the pixel embedding map along the boundaries.  Then, the features with the same/different labels in each patch are treated as the positive/negative samples, on which a classification procedure is performed to enhance the compactness of the features with the same label while distinguish the different semantic features.  Our boundary-perceiving loss 
%aims at enforcing the network to discriminate the features around image boundaries, it 
encourages the model to pay more attention to discriminate the features around boundaries, consequently, more boundary pixels could be identified.

Quantitative and qualitative results on BSDS500~\cite{BSDS500}, NYUv2~\cite{NYUV2}, ISIC-2017~\cite{ISIC2017} and ACDC~\cite{ACDC} datasets demonstrate that the proposed method achieves more outstanding performance against the state-of-the-art superpixel segmentation methods. In summary, we make the following contributions in this work:
\begin{itemize}
    \item We propose a novel AI module to directly capture the relation between the pixel and its surrounding grid cells, such a design builds a more consistent architecture with the target of superpixel segmentation.
    
    \item A boundary-perceiving loss is designed to facilitate network to accurately identify boundary pixels, which is achieved by discriminating the features with different semantic labels around boundaries.
    
    \item State-of-the-art performance on four benchmarks including two natural datasets BSDS500~\cite{BSDS500} and NYU~\cite{NYUV2} and two medical benchmarks ISIC-2017~\cite{ISIC2017} and ACDC~\cite{ACDC}. Besides, two stereo matching datasets, \ie SceneFlow~\cite{sceneFlow} and HR-VS~\cite{HRVS}, are further included to perform the downstream disparity matching task.  
\end{itemize}

This manuscript makes following improvements on the basis of our conference version AINet~\cite{ainet}:We extend our AINet from a single implantation scheme to a hierarchical one (named AINet+), making the network better capture the pixel-superpixel relation, and promoting our performance one step further.
To further verify the superiority of our methods, we additionally perform extensive experiments on two medical  benchmarks, \ie, ISIC-2017~\cite{ISIC2017}, and ACDC~\cite{ACDC}. Our AINet+ could outperform all of the competitive methods on these four datasets. 
What’s more, to study the benefits of our AINet+ to the downsteam task, we conduct experiments on the stereo matching on two popular benchmarks, SceneFlow~\cite{sceneFlow} and  HR-VS~\cite{HRVS}. By equipping our AINet+, the performance of stereo matching model could get improved as expected.
%\textbf{(1)} We introduce one more AI module at lower resolution layer to perform a hierarchical association implantation, which promotes our method to the new state-of-the-art on four benchmarks.
%\textbf{(2)} More datasets are included to evaluate the performance, we additionally introduce two medical benchmarks to validate the superiority of our method.
%\textbf{(3)} We further conduct experiments on stereo match task to evaluate the application value of the proposed method.

In the following, we will first introduce the existing works related to this paper in section~\ref{relatedWork}. Section~\ref{Pre} would give some prior knowledge to ease the understanding of this paper. In section~\ref{method}, we would elaborate the details of our proposed method. Experimental results on four benchmarks are presented in section~\ref{experiment}. Section~\ref{conclusion} gives the conclusions.

%------------------------------------------------------------------------
\section{Related Work}
\label{relatedWork}
\yx{Superpixel segmentation seeks to group similar pixels for an image and is a fundamental problem in computer vision, extensive efforts have been dedicated to this task due its important  application for the downstream tasks. Hereinafter, we would revisit the existing works of superpixel segmentation and its application.} 

\begin{figure*}
\begin{center}
\includegraphics[width=5.6in,height=2.2in]{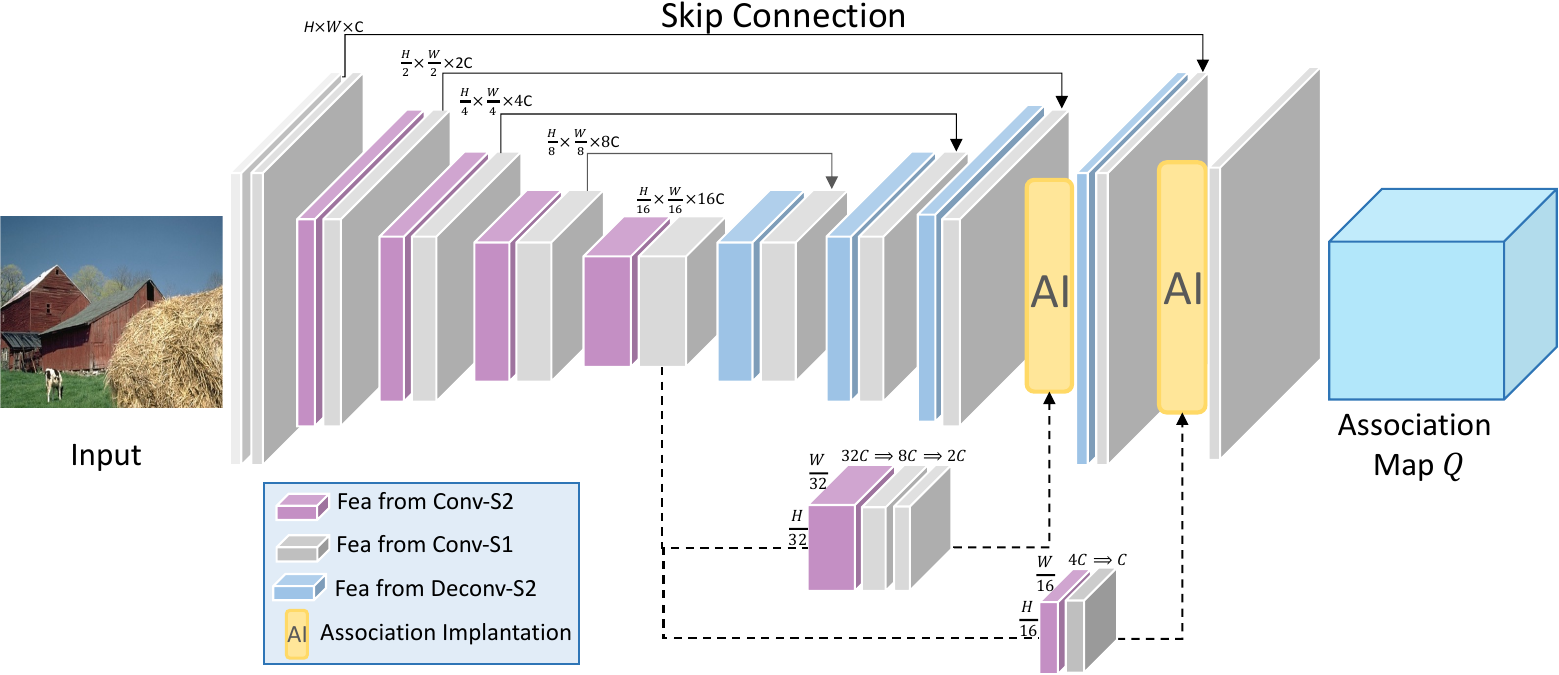}
\end{center}
\vspace{-0.4cm}
   \caption{\yx{The framework of our AINet+, where the conv/deconv means convolution/deconvolution, and conv-S\# means convolution with stride \#. The network takes an image as input and outputs the association map. Meanwhile, the superpixel embedding and pixel embedding are first obtained by the convolutions and then fed into the AI module to obtain the pixel-superpixel context. And the local patch loss is performed on the pixel-wise embeddings to boost the boundary precision. In AI module, the sampling interval is set to 16, and each block indicates a pixel or superpixel embedding.} }
\label{fig:workflow}
\vspace{-0.3cm}
\end{figure*}

\subsection{Superpixel Segmentation}  %Superpixel segmentation attempts to generate the image regions by grouping image pixels. Comparing to pixels, superpixel is a more natural representation of image data. 
Superpixel segmentation is a well-defined problem and has a long line of research~\cite{SP_evaluation,sp_add1,sp_add2,sp_add3,sp_add4,sp_add5}.  Traditional superpixel algorithms can be broadly classified into graph-based and clustering-based approaches.  Graph-based methods consider the pixels as nodes and the edges as strength of connectivity between adjacent pixels, respectively. Consequently, the superpixel segmentation could be formulated as a graph-partitioning problem. Wide-used algorithms, Felzenszwalb and Huttenlocher (FH)~\cite{FH} and the entropy rate superpixels (ERS)~\cite{ERS}, belong to this category. 
Liu~\etal~\cite{ERS} model the superpixel segmentation as a random walk on a graph, the authors utilize the entropy rate to make the superpixel compact and a balance term to encourage the clusters with the same size.  On the other hand, clustering-based approaches utilize classic clustering techniques like $k$-means to compute the connectivity between the anchor pixels and its neighbors. Well-known methods in this category include SLIC~\cite{SLIC}, LSC~\cite{LSC}, Manifold-SLIC~\cite{MSLIC} and SNIC~\cite{SNIC}. 
Achanta~\etal~\cite{SLIC} adapt the k-means algorithm to superpixel segmentation, the designed SLIC method achieves good performance and efficiency simultaneously. 
%Li~\etal~\cite{CSS} attempt to reduce the computation of superpixel segmentaion and propose a cluster sensing superpixel method, the proposed method is time-efficient. SNICPOLY~\cite{SNICPoly} extend the classic SNIC method and develop a more efficient approach.
Inspired by the success of deep learning techniques, recently, researchers attempt to utilize the deep neural network to learn the membership of each pixel to its surrounding grid cells.
Tu \etal~\cite{SEAL} attempt to learn deep features to benefit the superpixel segmentation, however, the overall architecture is still not differentiable. 
Jampani \etal~\cite{SSN} develop the first differentiable deep network motivated by the classic SLIC method, and Yang \etal~\cite{SCN} further simplify the framework and contribute a more efficient model. Zhu~\etal~\cite{SPcvpr2021} attempt to learn the superpixels in an unsupervised manner and propose to perform superpixel segmentaion by a lifelong clustering problem. 

\iffalse
\begin{figure}[t]
\setlength{\abovecaptionskip}{0pt} 
\setlength{\belowcaptionskip}{0pt} 
\begin{center}
    \includegraphics[width=2in]{first_shown-crop.pdf}
    %\vspace{0.02cm}
\end{center}
\vspace{-0.2cm}
\caption{For each pixel in image, only the 9 neighbors is considered for assignment.\ycnote{better to put this figure in the first page and make a comparison between ours and others.}}
\label{first_shown}
\end{figure}
\fi
%-------------------------------------------------------------------------

%The input image is first encoded to obtain to superpixel embeddings, several convolution \& pooling layers are followed to obtain the pixel-wise embedding. Then, the superpixel embedding and the pixel embedding are fed into the AI module to obtain the pixel-superpixel context. Finaly, a convolution is conducted to predict the association map.

\subsection{Application of Superpixel} The pre-computed superpixel segmentation could be viewed as a type of weak label or prior knowledge to benefit many downstream tasks like image \& video segmentaion~\cite{wei2,wei7,wei9,hrnet,seg}, object detection~\cite{qian1,qian2}, image super-resolution~\cite{iccv231} and so on. The superpixels could be integrated into deep learning pipeline to provide guidance so that some important image properties (e.g., boundaries) could be better preserved~\cite{app_add1,app_add2,app_add3,app_add4,app_add5,eccv241,cvpr24}. For example, KwaK \etal~\cite{SP_Pool1} utilize the superpixel segmentation to perform a region-wise pooling to make the pooled feature have better semantic compactness. In ~\cite{medical_sp}, Cheng \etal consider the superpixel as pseudo label and attempt to boost the image segmentation by identifying more semantic boundary. Besides benefiting the image segmentation or feature pooling, superpixel also provides flexible ways to represent the image data. He \etal~\cite{spcnn} convert 2D images patterns into 1D sequential representation, such a novel representation allows the deep network to explore the long-range context of the image.
Gadde \etal~\cite{sp_pool2} propose a bilateral inception module that propagates information between superpixels, the authors target on capturing more structure information with the guidance of superpixel. 
Liu \etal~\cite{DEL} learn the similarity between different superpixels, the developed framework could produce different grained segmentation regions by merging the superpixels according to the learned superpixel similarity. 
In~\cite{sp_attack}, Dong \etal utilize the superpixel to guide the attention procedure and develop a robust adversarial attack framework. 
Besides images, superpixel could also benefit the video-related tasks. For example, Wang~\etal~\cite{sp_app1} treat the superpixel as a prior knowledge and integrate it into a video segmentation framework, the author use the superpixel to perform spatial and temporal pooling, which could provide good guidance for the video segmentation. 

%In~\cite{sp_app2}, Li~\etal propagate the region information on multi-layer superpixels, which works together with a random refinement to generate candidate 3D label sets for each pixels. } 
%Gangapure~\etal~\cite{sp_app3} develop a superpixel-based causal multisensor video fusion and design a superpixel level spatial and temporal saliency models.

\section{Preliminaries}
\label{Pre}
Before delving into the details of our method, we first introduce the framework of deep-learning based superpixel segmentation, which is also the fundamental theory of this paper.  
As illustrated in Fig.~\ref{first_shown}, the image $I$ is  partitioned into blocks using a regular size grid, and the grid cell is regarded as the initial superpixel seed. For each pixel $p$ in image $I$, the superpixel segmentation aims at finding a mapping that assigns each pixel to one of its surrounding grids, \ie 9 neighbors, just as shown in Fig.~\ref{first_shown}. Mathematically, deep-learning based method feeds the image $I\in \mathcal{R}^{H\times W\times 3}$ to convolution neural network and output an association map $Q\in \mathcal{R}^{H\times W\times 9}$, which indicates the probability of each pixel to its neighbor grids~\cite{SSN,SCN}.  Since there is no ground-truth for such an output, the supervision for network training is performed in an indirect fashion: the predicted association map $Q$ serves as the intermediate variable to reconstruct the pixel-wise property $l(p)$ like semantic label, position vector, and so on. Consequently, there are two critical steps in training stage.

\textbf{Step1:} Estimate the superpixel property from the surrounding pixels:
\begin{equation}
\label{step1}
    h(s) = \frac{\sum_{p:s\in N_p}{l(p)\cdot q(p,s)}}{\sum_{p:s\in N_p}q(p,s)}.
\end{equation}

\textbf{Step2:} Reconstruct the pixel property according to the superpixel neighbors:
\begin{equation}
\label{step2}
    l^{'}(p) = \sum_{s\in N_p} h(s)\cdot q(p,s),
\end{equation}
where the $N_p$ is the set of adjacent superpixels of $p$, $q(p,s)$ indicates the probability of pixel $p$  assigned to superpixel $s$. 
The output $h(s)$ in step 1 is an average property for the initial superpixel $s$, which is estimated using the pixel association $q(p,s)$ to weight the pixel properties $l(p)$, the denominator $\sum_{p:s\in N_p}q(p,s)$ is used for normalization.  In the next step, the pixel property $l^{'}(p)$ is reconstructed in return using the association $q(p,s)$ to weighted sum the superpixel property estimation $h(s)$.  

\begin{figure}[t]
\setlength{\abovecaptionskip}{0pt} 
\setlength{\belowcaptionskip}{0pt} 
\begin{center}
    \includegraphics[width=3.5in]{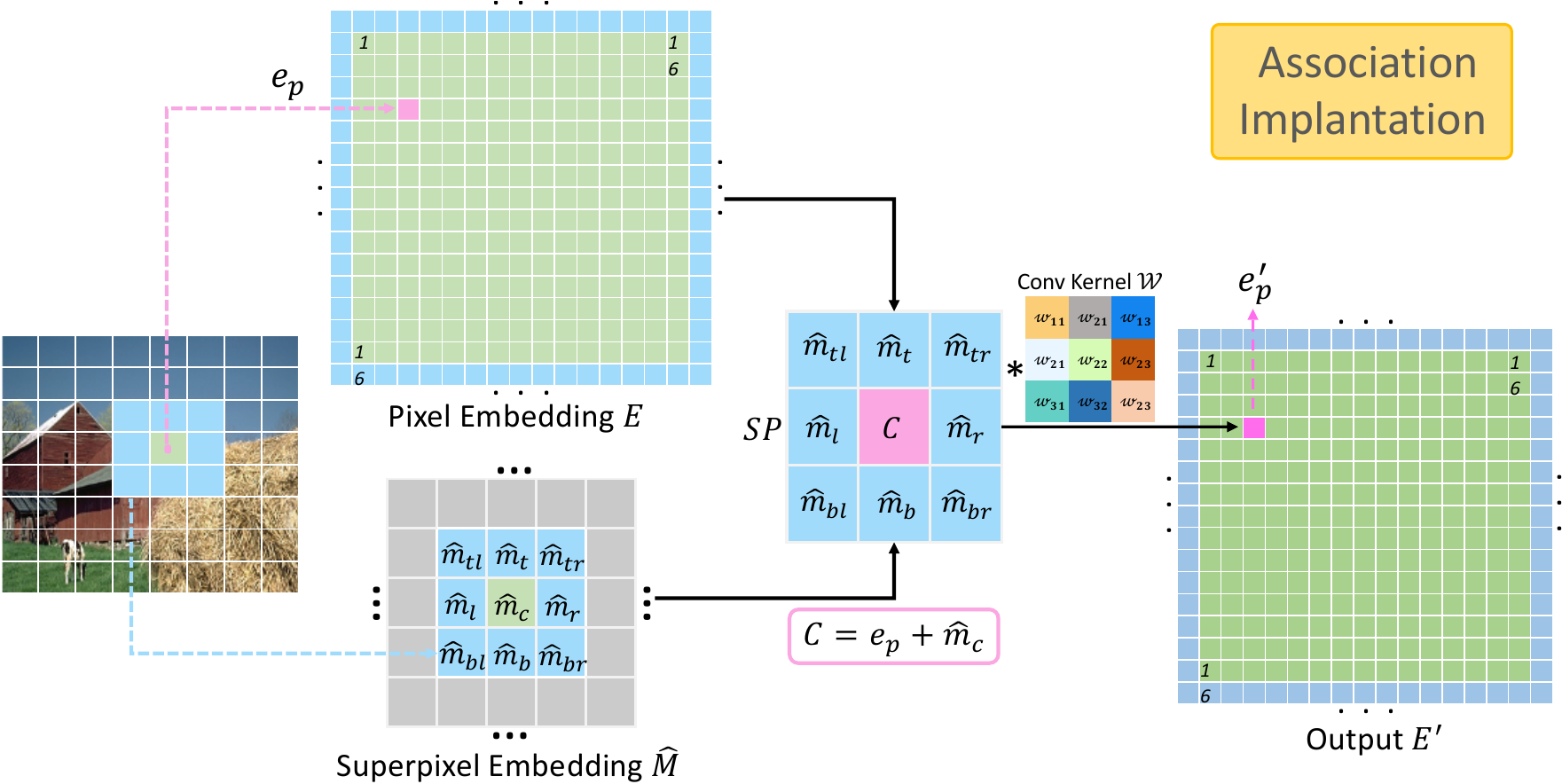}
    %\vspace{0.02cm}
\end{center}
%\vspace{-0.2cm}
\caption{Illustration of our Association Implantation Module (AI). We directly implant the surrounding grid features to the central pixels, such a design could allow the network to explicitly harvest the pixel-grid relation, which is the context exactly required by the superpixel segmentation.  
}
\label{aimodule}
\vspace{-0.2cm}
\end{figure}

Thus, the training loss is to optimize the distance between the ground-truth property and the reconstructed one:
\begin{equation}
\label{eq1}
    \mathcal{\bm{L}}(Q) = \sum_{p} dist(l(p),l^{'}(p))).
\end{equation}

\yx{Following Yang's practice~\cite{SCN}}, the properties of pixel in this paper include the semantic label and the position vector, \ie, two-dimension spatial coordinates, which are optimized by the cross-entropy loss and the $\mathcal{L}_2$ reconstruction loss, respectively.

%------------------------------------------------------------------------
\section{Methodology}
\label{method}
An overview of our proposed AINet+ is shown in Fig.~\ref{fig:workflow}. In general, the overall architecture is an encoder-decoder style paradigm, the encoder module compresses the input image and outputs a feature map called superpixel embedding, whose pixels exactly encode the features of grid cells. Subsequently, the superpixel embedding is further fed into the decoder module to produce the association map. Meanwhile, the superpixel embedding and the pixel embedding in decoding stage are integrated to perform the association implantation, we also perform our association implantation on the second-last decoding layers to hierarchically capture the pixel-superpixel relation.  The boundary-perceiving loss acts on the pixel embedding to help discriminate the pixels around boundary in hidden feature level. Hereinafter, we will first elaborate the details of our proposed AI module in subsection~\ref{AImodule}, in subsection~\ref{hai}, the hierarchical association implantation will be presented, and the designed boundary-perceiving loss would be introduced in subsection~\ref{BPL}.

\subsection{Association Implantation Module}
\label{AImodule}
%\small{As introduced in Section~\ref{Pre}, the superpixel segmentation is achieved by assigning each pixel to its 9 neighbor cells. In such a procedure, the information propagation between the pixel and its neighbors is much important to capture their relation. However, existing methods simply predict the association map by recurrently employing the convolution/deconvolution operations, which is not effective to capture the accurate relation between the pixel and its surrounding cells.
%no special design is proposed to transfer the knowledge between pixel and its surrounding cells. 
%To remedy this weakness, this work proposes an association implantation module to perform a direct interaction between the pixel and its neighbor cells. We first obtain the embeddings of superpixels and pixels by the convolution network, then, for each pixel embedding, its corresponding neighbor superpixel features are picked and inserted to its surrounding, finally, a convolution with kernel size $3\times 3$ is conducted on the expanded pixel embedding to achieve the knowledge transferring.}

To enable the network to explicitly perceive the relation between each pixel and its surrounding grid cells, this work proposes an association implantation module to perform a direct interaction between the pixel and its neighbor grids. As shown in Fig.~\ref{aimodule}, we first obtain the embeddings of superpixels and pixels by the convolution network. Then, for each pixel embedding, the corresponding neighbor superpixel features are picked and  \yx{implanted} to its surrounding. Finally, a convolution with kernel size $3\times 3$ is conducted on the expanded pixel embedding to achieve the knowledge propagation.

Formally, let $e_p\in\mathcal{R}^D$ be the embedding of pixel $p$ from the pixel embedding $E\in\mathcal{R}^{H\times W\times D}$, which is obtained by the deep neural network as shown in Fig.~\ref{fig:workflow}. To obtain the embeddings of the grid cells, \ie, superpixel embedding, we compress the input image by $\log_\text{2}S$ times using multiple convolutions and max-pooling operations, where $S$ is the \yx{sampling interval} for the grid cell. For example, if the sampling interval is 16, then, we downsample the image 4 times. This would result in a feature map $M\in\mathcal{R}^{h\times w\times D^{'}}$ whose pixels exactly encode the features of grid cells, where $h=H/S, \text{and } w=W/S$. To perform the implantation operation on the pixel embedding, we first adjust the channels of $M$ using two $3\times 3$ convolutions, producing a new map $\hat{M}\in\mathcal{R}^{H\times W\times D}$. Then, for the pixel $p$, we pick up its 9 adjacent superpixel embeddings from left to right and top down: $\{\hat{m}_{tl}, \hat{m}_{t}, \hat{m}_{tr}, \hat{m}_{l}, \hat{m}_{c}, \hat{m}_{r}, \hat{m}_{bl}, \hat{m}_{b},\hat{m}_{br}\}$ from $\hat{M}$. To allow the network could explicitly capture the relation between pixel $p$ and its neighbor grids,
%we first adjust the dimension of the superpixel embeddings using 2 convolution layers:
%\begin{equation}
%    \label{convs}
%    \hat{m}_{k} = \text{Conv}s(m_{k};\Theta), k\in\{tl,t,tr,l,c,r,bl,b,br\}.
%\end{equation}
 we directly implant the superpixel embeddings into the surrounding of the pixel $p$ to provide pixel-superpixel context:
\begin{equation}
\label{expanded_pixel}
SP=\left[
    \begin{matrix}
    \hat{m}_{tl} &\hat{m}_t & \hat{m}_{tr}\\
    \hat{m}_{l} & \hat{m}_{c} +e_p & \hat{m}_{r}\\
    \hat{m}_{bl} & \hat{m}_{b} & \hat{m}_{br}
    \end{matrix}
    \right].
\end{equation}
\iffalse
What's more, a greedy version of $SP$ could also be obtained by further including the neighbor pixels of $p$, which could enable the network to harvest the pixel-superpixel level context without losing the pixel-wise relation:
\begin{equation}
\label{expanded_pixel_new}
SP=\left[
    \begin{matrix}
    \hat{m}_{tl}+e_{tl} &\hat{m}_t+e_{t} & \hat{m}_{tr} + e_{tr}\\
    \hat{m}_{l} +e_{l}& \hat{m}_{c} +e_p & \hat{m}_{r}+e_{r}\\
    \hat{m}_{bl}+e_{bl} & \hat{m}_{b} + e_{b} & \hat{m}_{br} + e_{br}
    \end{matrix}
    \right],
\end{equation}
where $\{e_{tl}, e_{t}, e_{tr}, e_{l}, e_{c}, e_{r}, e_{bl}, e_{b}, e_{br}\}$ are the adjacent pixel embeddings of pixel $p$. 
\fi
It is worth noting that the pixels in the same initial grid would share the same surrounding superpixels, since they would degrade into one element in superpixel view. 
%where the sign of superpixel embedding remains unchanged to simplify the expression. 
We then adopt a $3\times 3$ convolution to adaptively distill information from the expanded window to benefit the subsequent association map inferring:
\begin{equation}
    \label{fuse_sp}
    e^{'}_p = \sum_{ij} SP_{ij}\times w_{ij} +b,
\end{equation}
where $w$ and $b$ are the convolution weight and bias, respectively.
We traverse all of the pixel embeddings in $E$ and apply the operations in Eq.~\ref{expanded_pixel}-~\ref{fuse_sp}, thus, we could obtain a new pixel embedding $E^{'}$ whose elements capture the pixel-superpixel level context. In the following, the feature map $E^{'}$ is fed through a convolution layer to predict the association map $Q$.

As shown in Eq.~\ref{expanded_pixel}-~\ref{fuse_sp}, our AI module directly places the neighbor grid embeddings in the surrounding of the pixel to provide the context required by superpixel segmentation, which is an intuitive and reasonable solution. Comparing to the existing methods that use the stacked convolutions to accumulate the pixel-wise relation, the pixel-superpixel context captured by our AI module is more in line with the target of superpixel segmentation.

\subsection{Towards Hierarchical Association Learning}
\label{hai}
%By feeding the pixel embeddings and its superpixel features, our AI module could capture the association in pixel level. 
Before predicting the membership of pixels, perceiving the pixel-grid relation in region level would ease the following fine-grained assignment of pixels, this could be validated by many progressive and hierarchical frameworks~\cite{pgan,hierar1,hierar2}. Motivated by this, we enhance our network with a hierarchical association implantation. As shown in Fig.~\ref{fig:workflow}, we further apply our association implantation operation at the previous layer $F$ of the pixel-embedding, the consecutive two AI modules enable the network to progressively harvest the pixel-grid relations. 

%To further boost the performance, we also attempt Inspired by the success of the hierarchical structure,   

The pixels in $F$ encode a small $2\times 2$ regions of the input image, to harvest its superpixel embedding $M_\downarrow$, we further down-sample the feature map $M$ by a $3\times 3$ convolution with stride $2$, such that each pixel in $M_\downarrow$ captures the features of a $2S\times 2S$ region of the input image.  We next acquire $\hat{M}_\downarrow$ by adjusting the channel dimension of $M_\downarrow$ and perform the association implantation using $F$ and $\hat{M}_\downarrow$. Different from the AI operation on the full resolution level, this association implantation is performed on the region level, which works together with the following AI module to form a hierarchical structure and  enables the network to capture the pixel-superpixel relation in a coarse-to-fine manner.

%\ycnote{Since AI module is the main contribution of this work, better to add more content.}
%On the other hand, it also provides broader contextual clues to help learn more effective features. Finally, we adopt a convolution with kernel size $3\times 3$ to adaptively distill information from the expanded window to benefit the subsequent association map prediction.  

%Specially, for each pixel embedding, we expand it to a $3\times 3$ window, thus, the original embedding map $E$ is enlarged $3\times 3$ times. Then a convolution with kernel size 3 and stride 3 is applied to get a feature map %$\mathcal{M}$ whose shape would be $H\times W\times D$:
% \begin{equation}
%     \mathcal{M}_i = Conv(SP,ks=3,std=3; \theta).
% \end{equation}

\subsection{Boundary-Perceiving Loss}
\label{BPL}
\begin{figure}[t]
\setlength{\abovecaptionskip}{0pt} 
\setlength{\belowcaptionskip}{0pt} 
\begin{center}
    \includegraphics[width=3.5in]{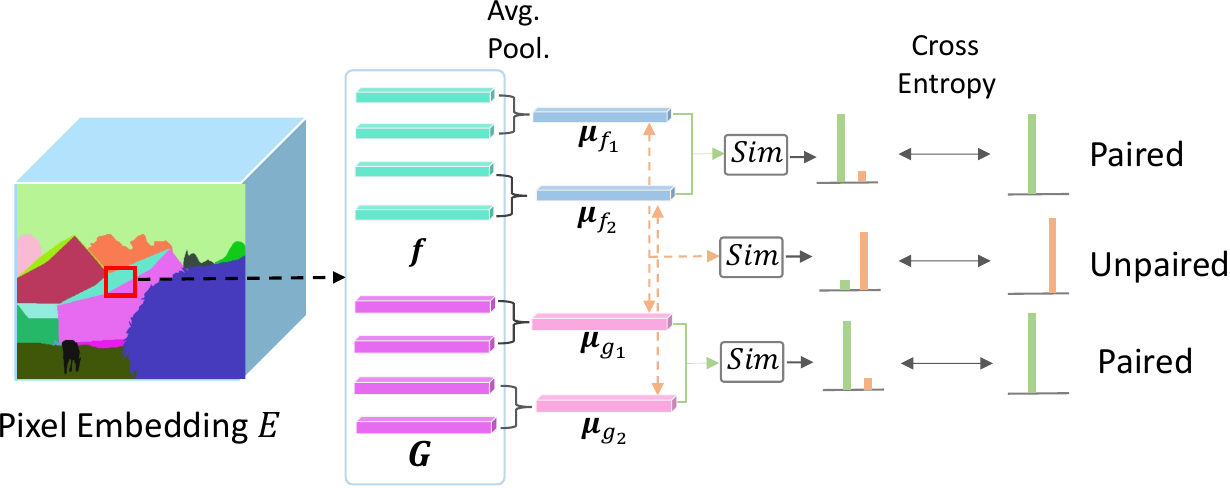}
    %\vspace{0.02cm}
\end{center}
%\vspace{-0.2cm}
\caption{Illustration of our Boundary-perceiving Loss (BPL). By performing a classification procedure on the boundary patches, we targets at discriminating the boundary pixels in hidden feature level, which is achieved by enforcing the pixels with the same semantic label to be close while the different ones to be far away.  
}
\label{bpl}
\vspace{-0.2cm}
\end{figure}

%\ycnote{local patch loss sounds not good. bilateral? contrastive? boundary-perceiving?}
%\ycnote{Reduce the content}
%The pixels in the same superpixel should share compact semantics, this is especially important when grouping the pixels around boundaries. To make the pixels on different sides of the boundary be appropriately assigned, we attempt to boost the discrimination of the features around boundaries to benefit the following pixel assignment, consequently, a local patch loss is proposed.  
Our boundary-perceiving loss is proposed to help the network appropriately assign the pixels around boundaries. As shown in Fig.~\ref{bpl}, we first  sample a series of patches with a certain size (5$\times$5, for example) around boundaries in the pixel embedding map, and then a classification procedure is conducted to improve the discrimination of the different semantic features.
%enforce the feature embeddings in these local patches to be discriminative enough to ease the following pixel grouping.

Formally, let $E\in \mathcal{R}^{H\times W\times D}$ be the pixel-wise embedding map, since the ground-truth label is available during training stage, we could sample a local patch $B\in \mathcal{R}^{K\times K\times D}$ surrounding a boundary pixel from $E$. For the sake of simplification, the patch $B$ only covers the pixels from two different semantic regions, that is, $B=\{f_1,\cdots, f_m, g_1, \cdots, g_n\}$, 
%where superscript and subscript indicates the categories and item indices, respectively, 
where $f, g\in \mathcal{R}^{D}, m+n=K^2$. 
Intuitively, we attempt to make the features in the same categories be closer, while the embeddings from different labels should be far away from each other. To this end, we evenly partition the features in the same categories into two groups, $\bm{f}^1, \bm{f}^2, \bm{g}^1, \bm{g}^2$, and employ a classification-based loss to enhance the discrimination of the features:
\begin{equation}
\begin{aligned}
    \mathcal{\bm{L}}_{B} &= -\frac{1}{2}(\log(sim(\bm{\mu}_{f1},\bm{\mu}_{f2})) + \log(1-sim(\bm{\mu}_{f1},\bm{\mu}_{g1}))) \\
    &- \frac{1}{2}( \log(sim(\bm{\mu}_{g1}, \bm{\mu}_{g2}))+ \log(1-sim(\bm{\mu}_{f2},\bm{\mu}_{g2}))),
\end{aligned}
\end{equation}
where the $\bm{\mu}_{f1}$ is the average representation for $\bm{f}^1$, and the function $sim(\cdot,\cdot)$ is the similarity measure for two vectors:
\begin{gather}
    \bm{\mu}_{f1} = \frac{1}{|\bm{f}^1|}\sum_{f\in \bm{f}^1} f,\\
    sim(f,g) = \frac{2}{1+\exp(||f-g||_1)},
\end{gather}

Taking all of the sampled patches $\mathcal{\bm{B}}$ into consideration, our full boundary-perceiving loss is formulated as follow:
\begin{equation}
    \mathcal{\bm{L}}_{\mathcal{B}} = \frac{1}{|\mathcal{\bm{B}}|}\sum_{B\in\mathcal{\bm{B}}} {\mathcal{L}_B}.
\end{equation}
%The proposed local patch loss focuses on the regions around boundaries and help the network identify the boundary pixel more accurately by highlighting the features 

\iffalse
On the other hand, another thought to achieve our idea is to directly optimize the distance between features in the local patch, \ie, enhance the connection in the same category while differentiate the features from different categories. Motivated by this, we propose another patch loss as follow:
\begin{equation}
    \mathcal{\bm{L}}^{'}_B = -\frac{||\bm{\mu}_f-\bm{\mu_g}||^2_2}{\bm{S}_f^2 + \bm{S}_g^2},
\end{equation}
where $\bm{\mu}_f$ and $\bm{S}_f$ are the average representation and compactness measure for features $\{f_i\}_{i=1}^m$:
\begin{gather}
    \bm{\mu}_{f} = \frac{1}{|\bm{f}|}\sum_{f\in \bm{f}} f,\\
    \bm{S}_f = \sum_{f\in \bm{f}} ||f-\bm{\mu_f}||_2^2,
\end{gather}
Subsequently, the full loss reads:
\begin{equation}
    \bm{L}_{\mathcal{B}}^{'} = \frac{1}{|\mathcal{\bm{B}}|}\sum_{B\in\mathcal{\bm{B}}} \mathcal{L}_B^{'}.
\end{equation}
\fi

%With the proposed local patch loss, the network could better identify the boundary pixels in the feature level and accurately group them in the following steps, consequently, the boundary precision could be improved.  
Overall, the full losses for our network training comprise three components, \ie, cross-entropy ($CE$) and $\mathcal{L}_2$ reconstruction losses for the semantic label and position vector according to the Eq.~\ref{eq1}, and our boundary-perceiving loss:
\begin{equation}
\label{total_loss}
    \mathcal{\bm{L}} = \sum_{p} CE(l^{'}_s(p), l_s(p)) + \lambda||p-p^{'}||_2^{2} + \alpha \mathcal{\bm{L}}_{\mathcal{B}}
\end{equation}
where $l^{'}_s(p)$ is the reconstructed semantic label from the predicted association map $Q$ and the ground-truth label $l_s(p)$ according to Eq.~\ref{step1}-~\ref{step2}, and $\lambda$, $\alpha$ are two trade-off weights.

\begin{figure}[t]
\setlength{\abovecaptionskip}{-0pt} 
\setlength{\belowcaptionskip}{-4pt} 
\begin{center}
    \subfigure[Patch shuffle]
    {  \begin{minipage}[t]{0.47\linewidth}
    \includegraphics[width=1.58in]{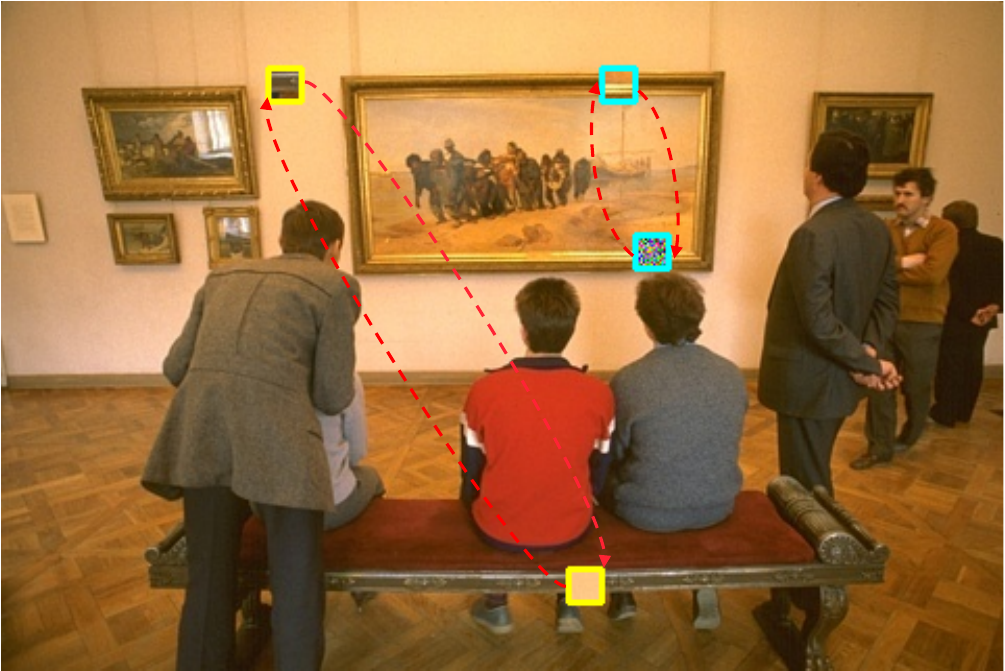}
    %\vspace{-2cm}
    %\includegraphics[width=1.52in]{image_edit_outpainting/visual_image/3_2_noRight.jpg}
    \end{minipage}}
    %\hspace{0.2cm}
    \subfigure[Random shift]{
    \begin{minipage}[t]{0.47\linewidth}
    \includegraphics[width=1.58in]{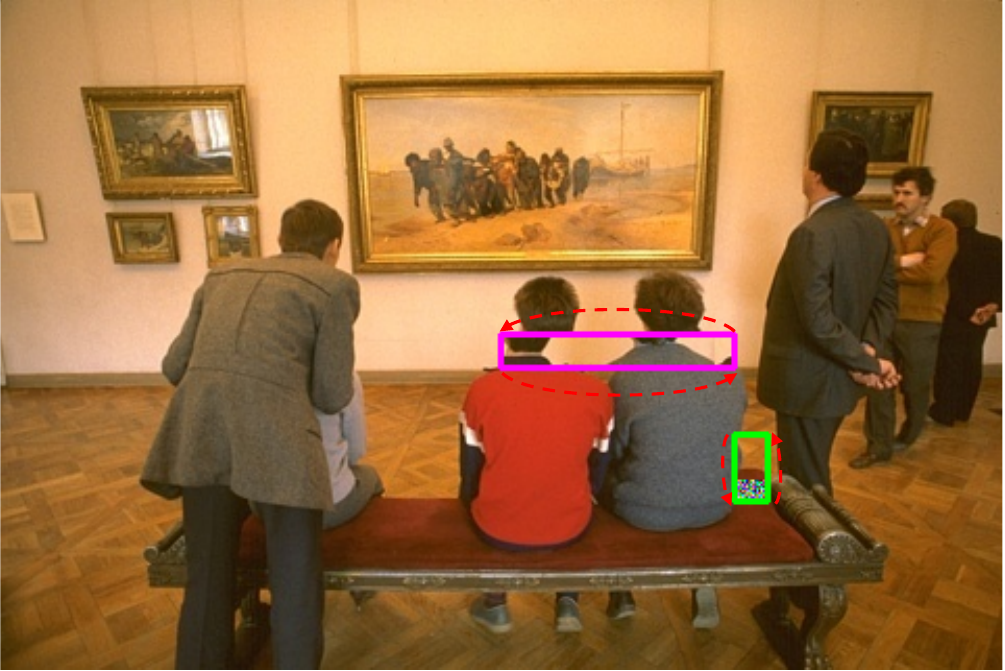}
    %vspace{-2cm}
    %\includegraphics[width=1.52in]{image_edit_outpainting/visual_image/3_2.jpg}
    \end{minipage}
    }
\end{center}
\vspace{-0.2cm}
\caption{The illustrations for our patch jitter augmentation, patch shuffle and random shift. Color frames indicate the changed regions.}
\label{vis_dataAug}
\end{figure}

\begin{figure*}[t]
\setlength{\abovecaptionskip}{-1pt} 
\setlength{\belowcaptionskip}{-7pt} 
\begin{center}
    \subfigure[\scriptsize BR-BP on BSDS500]{
    \begin{minipage}[t]{0.23\linewidth}
        \includegraphics[width=1.67in]{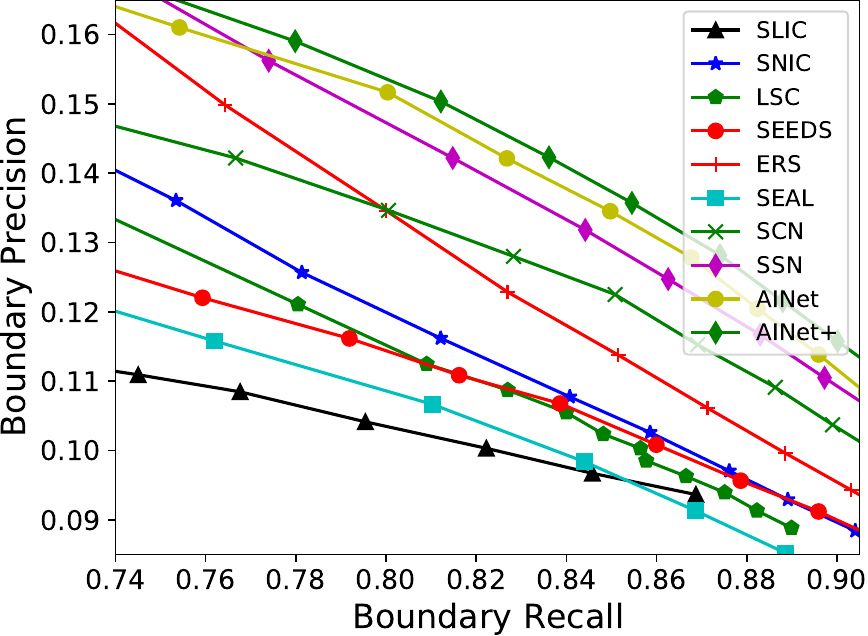}\\
    \vspace{-0.2cm}
    \end{minipage}
    }
    \hspace{0.02cm}
    \subfigure[\scriptsize ASA on BSDS500]{\begin{minipage}[t]{0.23\linewidth}
    \includegraphics[width=1.71in]{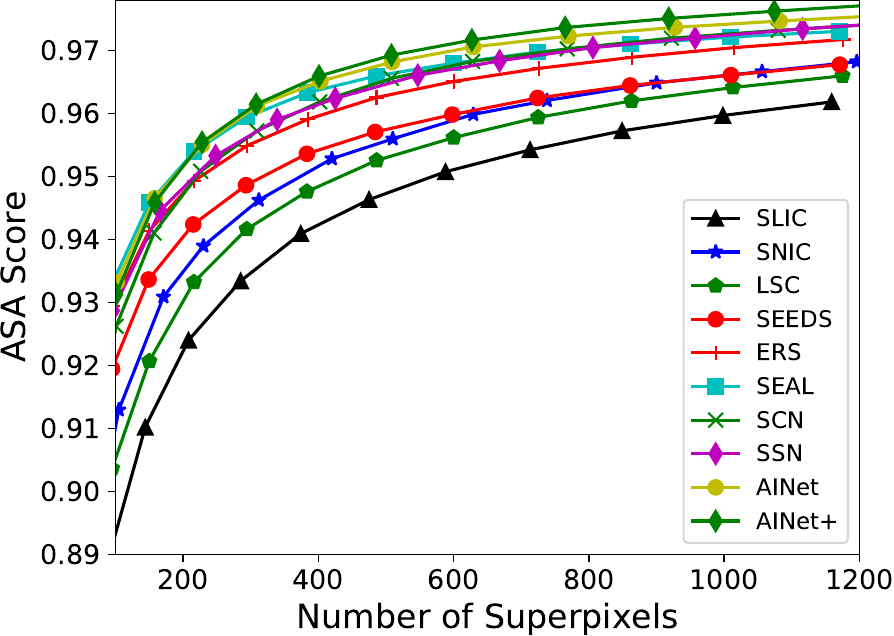}\\
    %\includegraphics[width=1.30in]{image_edit_outpainting/original_sketch_competing/yzx_528.jpg}\\
    %\vspace{0.01cm}
    %\includegraphics[width=1.30in]{image_edit_outpainting/original_sketch_competing/yzx_873.jpg}
    %\vspace{0.01cm}
    \vspace{-0.2cm}
    \end{minipage}}
    \hspace{0.02cm}
    \subfigure[\scriptsize BR-BP on NYUv2]{
    \begin{minipage}[t]{0.23\linewidth}
        \includegraphics[width=1.67in]{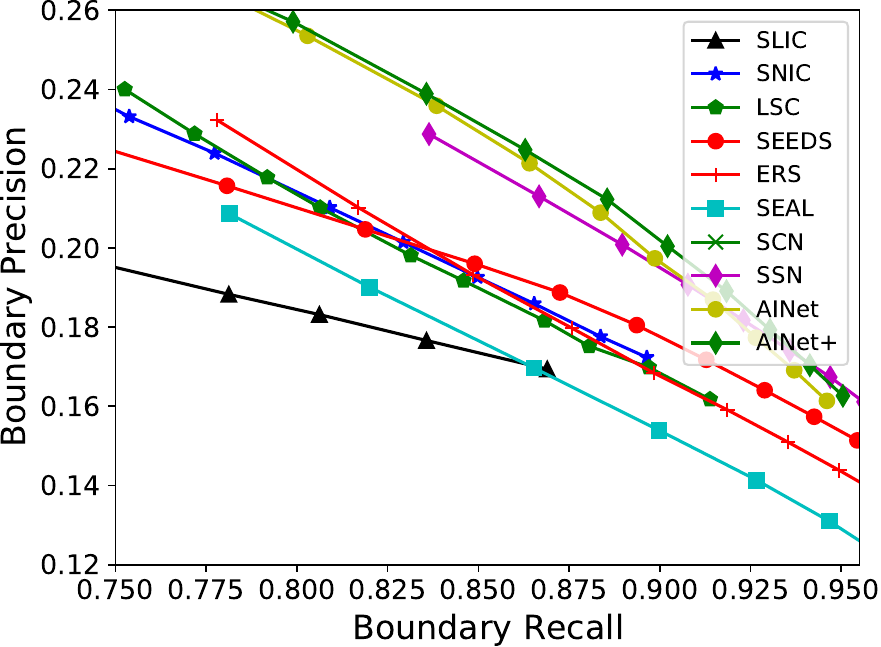}\\
    \vspace{-0.2cm}
    \end{minipage}
    }
    %\hspace{0.01cm}
    \subfigure[\scriptsize ASA on NYUv2]{
    \begin{minipage}[t]{0.23\linewidth}
    \includegraphics[width=1.67in]{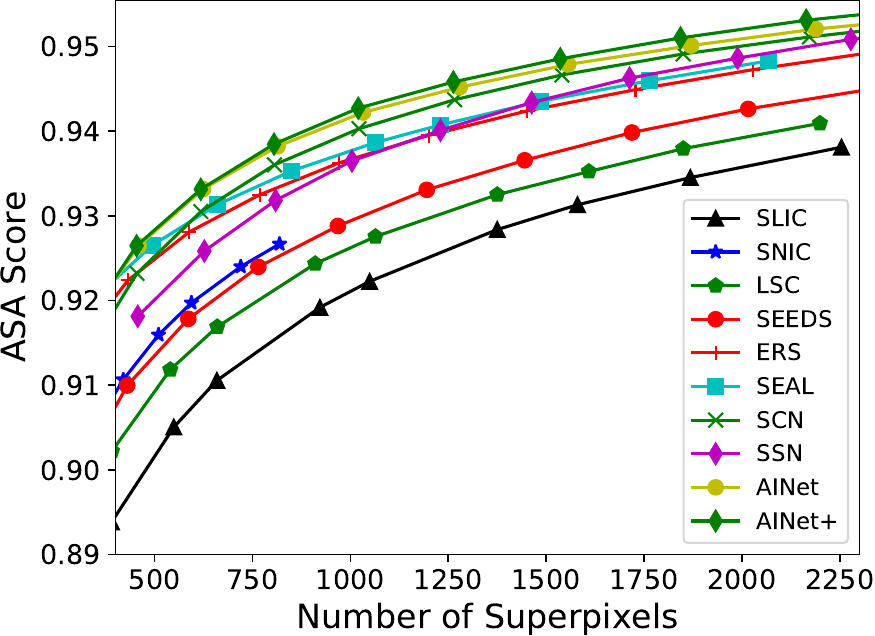}\\
    %\includegraphics[width=1.30in]{image_edit_outpainting/original_sketch_competing/our_528.jpg}\\
    %\vspace{0.02cm}
    %\includegraphics[width=1.30in]{image_edit_outpainting/original_sketch_competing/our_873.jpg}
    \vspace{-0.2cm}
    \end{minipage}
    }
\end{center}
\vspace{-0.3cm}
 \caption{Performance comparison on datasets BSDS500 and NYUv2.}
\label{main_perform}
\end{figure*}

\begin{figure*}[t]
\setlength{\belowcaptionskip}{-4pt} 
\begin{center}
    \subfigure[Inputs]{
    \begin{minipage}[t]{0.14\linewidth}
    \includegraphics[width=1.12in]{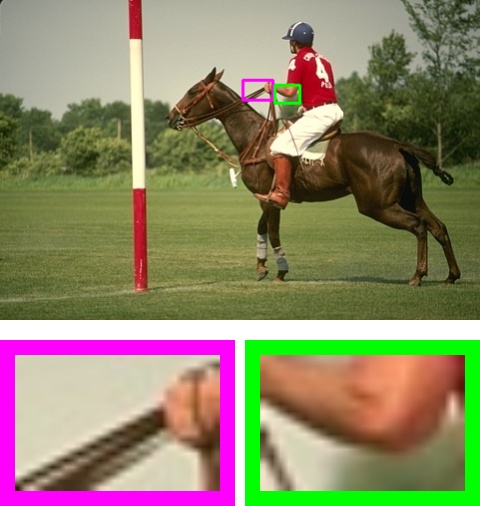}\\
    \vspace{-0.3cm}
    \includegraphics[width=1.12in]{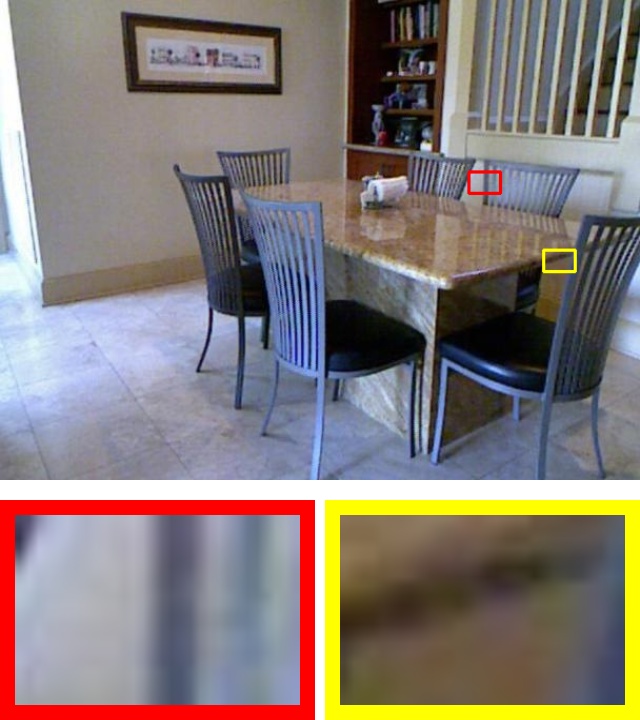}\\
     %\vspace{-0.4cm}
    %\includegraphics[width=1.35in]{tcyb/original_sketch_competing/pix2pix_528.jpg}\\
    \vspace{-0.3cm}
    \end{minipage}
    }
   \hspace{0.02cm}
    \subfigure[GT label]{
    \begin{minipage}[t]{0.15\linewidth}
    \includegraphics[width=1.12in]{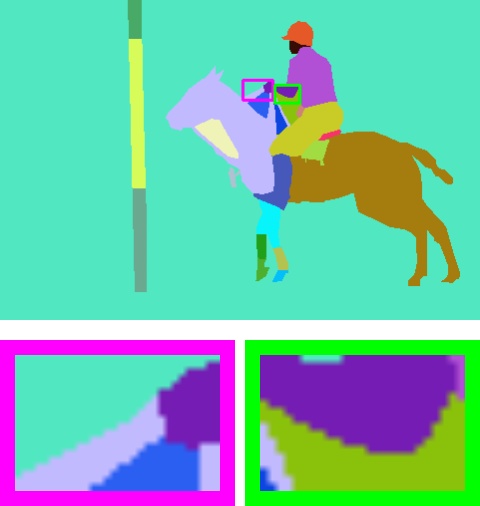}\\
     \vspace{-0.3cm}
    %\includegraphics[width=1.35in]{tcyb/original_sketch_competing/yzx_873.jpg}\\
     %\vspace{-0.4cm}
    \includegraphics[width=1.12in]{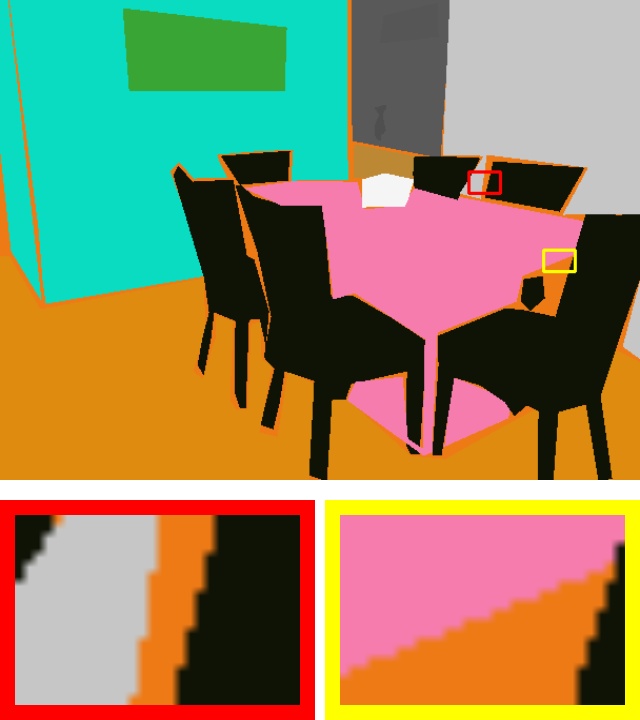}\\
     \vspace{-0.3cm}
    \end{minipage}}
    \hspace{-0.06cm}
    \subfigure[SCN~\cite{SCN}]{
    \begin{minipage}[t]{0.15\linewidth}
    \includegraphics[width=1.12in]{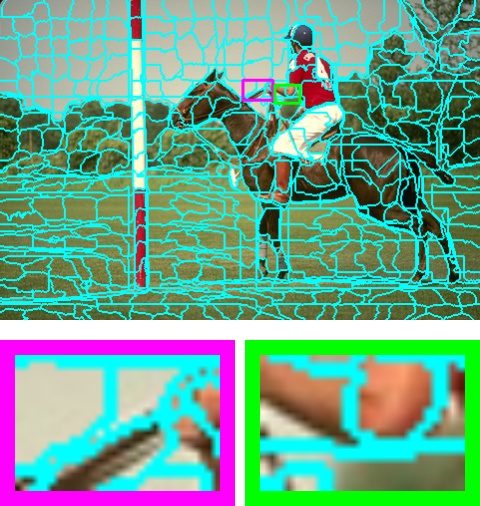}\\
     \vspace{-0.3cm}
    %\includegraphics[width=1.35in]{tcyb/original_sketch_competing/our_873.jpg}\\
    % \vspace{-0.4cm}
    \includegraphics[width=1.12in]{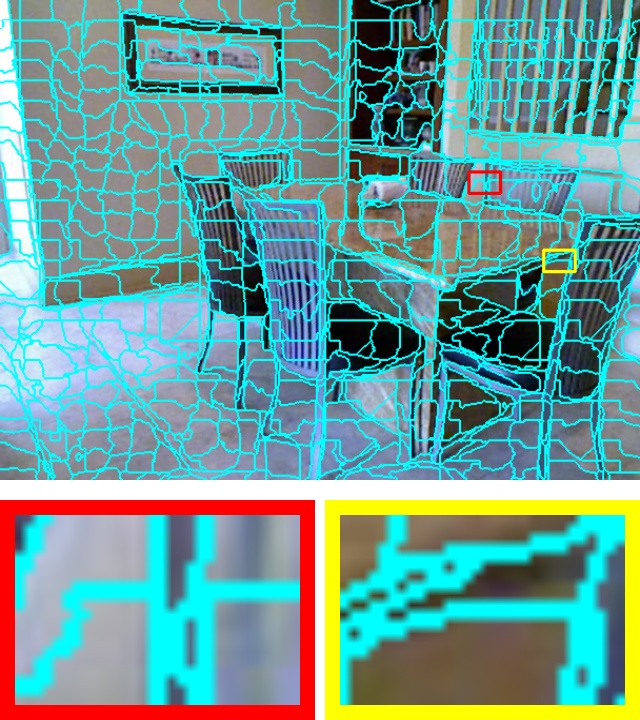}\\
     \vspace{-0.3cm}
    \end{minipage}
    }
    \hspace{-0.18cm}
    \subfigure[SSN~\cite{SSN}]{
    \begin{minipage}[t]{0.15\linewidth}
    \includegraphics[width=1.12in]{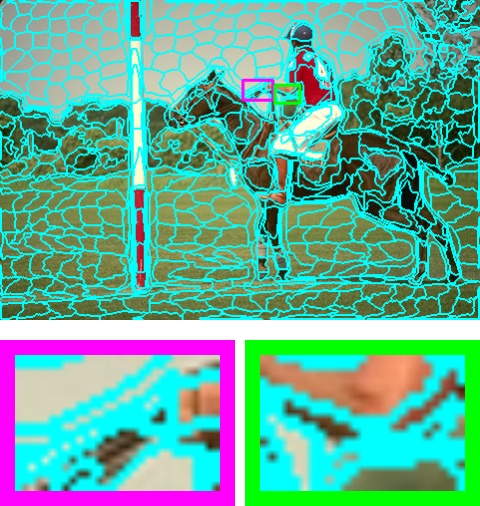}\\
    \vspace{-0.3cm}
    %\includegraphics[width=1.35in]{tcyb/original_sketch_competing/ori_873.jpg}\\
    % \vspace{-0.4cm}
    \includegraphics[width=1.12in]{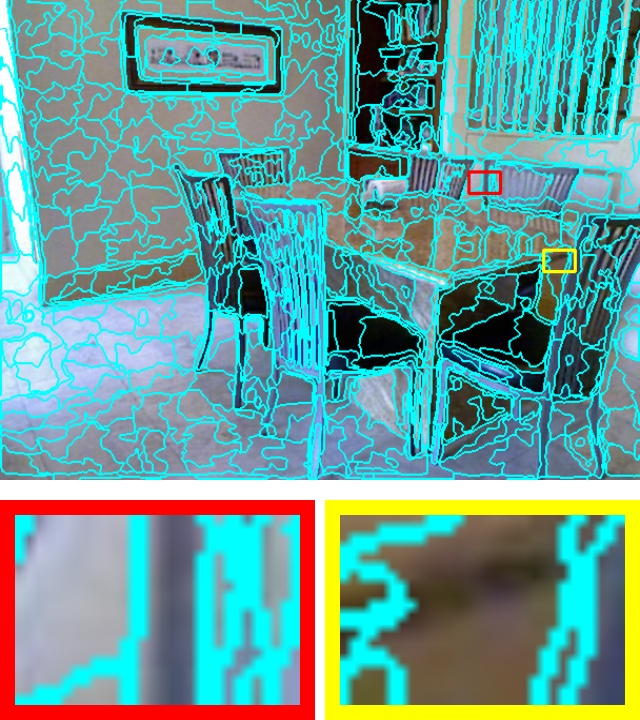}\\
     \vspace{-0.3cm}
    \end{minipage}
    }
     \hspace{-0.18cm}
    \subfigure[AINet]{
    \begin{minipage}[t]{0.15\linewidth}
    \includegraphics[width=1.12in]{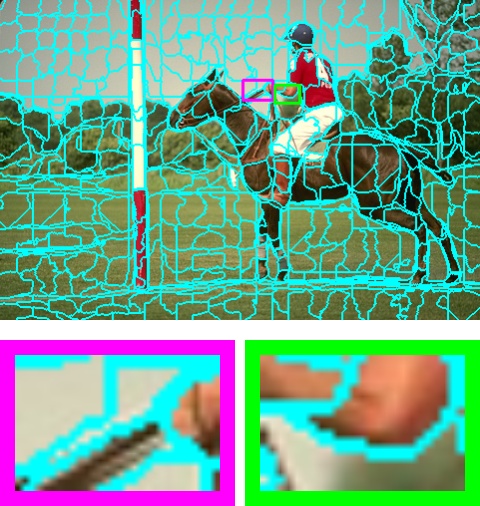}\\
    \vspace{-0.3cm}
    %\includegraphics[width=1.35in]{tcyb/original_sketch_competing/ori_873.jpg}\\
    % \vspace{-0.4cm}
    \includegraphics[width=1.12in]{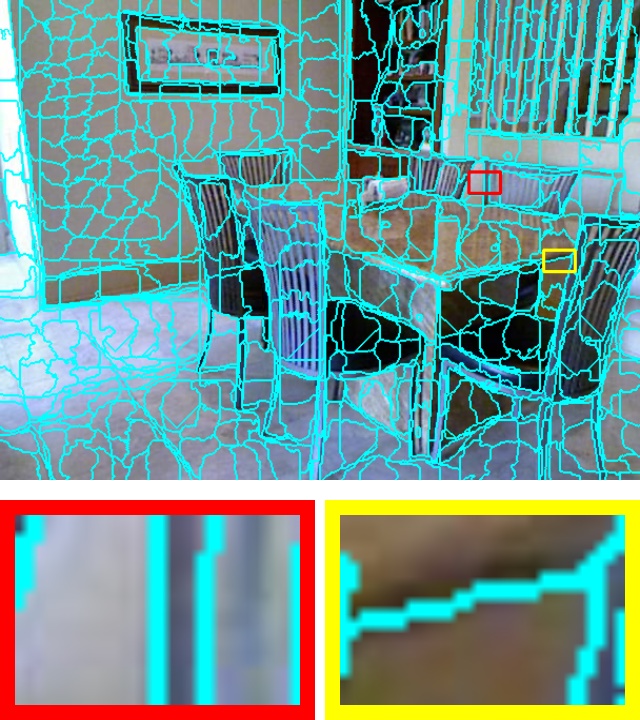}\\
     \vspace{-0.3cm}
    \end{minipage}
    }
    \hspace{-0.18cm}
    \subfigure[AINet+]{
    \begin{minipage}[t]{0.15\linewidth}
    \includegraphics[width=1.12in]{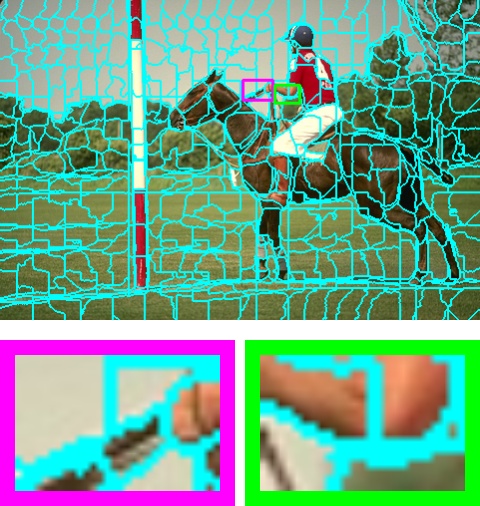}\\
     \vspace{-0.3cm}
    %\includegraphics[width=1.35in]{tcyb/original_sketch_competing/BDIE_873.png}\\
    % \vspace{-0.4cm}
    \includegraphics[width=1.12in]{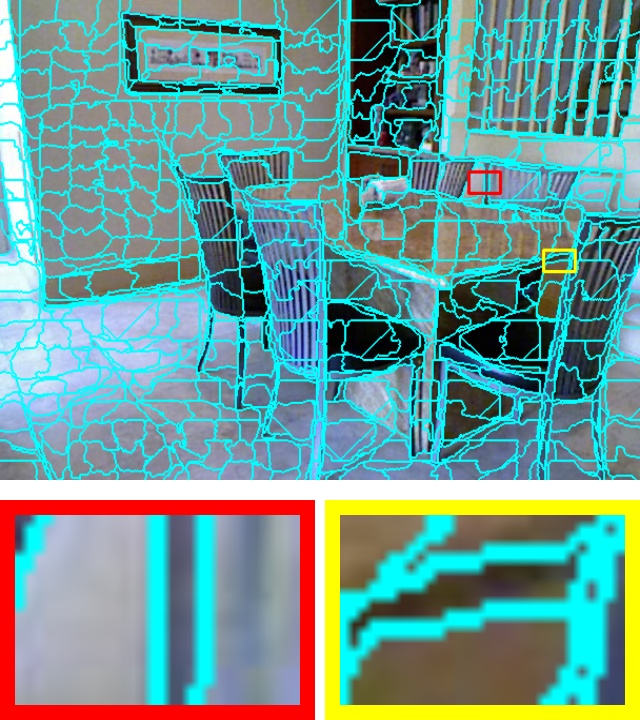}\\
     \vspace{-0.3cm}
    \end{minipage}
    }
\end{center}
\vspace{-0.6cm}
 \caption{Qualitative results of four SOTA superpixel methods, SEAL, SCN, SSN, and our AINet. The top row exhibits the results from BSDS500 dataset, while the bottom row shows the superpixels on NYUv2 dataset.}
 %\vspace{-0.2cm}
\label{main_visual}
\end{figure*}

\section{Experiments}
\label{experiment}
\noindent{\textbf{Datasets.}} We conduct experiments on four public benchmarks, \textbf{BSDS500}~\cite{BSDS500}, \textbf{NYUv2}~\cite{NYUV2}, \textbf{ISIC-2017}~\cite{ISIC2017}, and \textbf{ACDC}~\cite{ACDC} to evaluate the effectiveness of our method. BSDS500 comprises 200 training, 100 validation and 200 test images, and each image is annotated by multiple semantic labels from different experts. To make a fair comparison, we follow previous works~\cite{SCN,SSN,SEAL} and treat each annotation as an individual sample. Consequently, 1,087 training, 546 validation samples and 1,063 testing samples could be obtained. NYUv2 is an indoor scene understanding dataset and contains 1,449 images with object instance labels. To evaluate the superpixel methods, Stutz \etal~\cite{SP_evaluation} remove the unlabelled regions near the boundary and collect a subset of 400 test images with size 608$\times$448 for superpixel evaluation.  To futher validate the effectiveness of our proposed model, we also additionally introduce two medical datasets, \textbf{ISIC-2017}~\cite{ISIC2017} and \textbf{ACDC}~\cite{ACDC}, for performance evaluation. The ISIC-2017 dataset contains skin lesion images and the corresponding manually annotated lesion dilineations belonging to four diagnoses: melanoma, seborrheic, keratosis, and benign nevi. Following the official split, we use 2,000, 150, and 600 for training, validation and testing, respectively. ACDC contains 100 cine magnetic resonance (MR) exams covering well defined pathologies: dilated cardiomyopathy, hypertrophic cardiomyopathy, myocardial infarction with altered left ventricular ejection fraction and abnormal right ventricle, and also include normal subjects. Each exam only contains acquisitions at the diastolic and systolic phases. We use 75 exams for training and 10 exams for validation, the left 15 exams is used for testing.  

Following Yang's practice~\cite{SCN}, we conduct a standard train and test pipeline on the BSDS500 dataset. On the subject of NYUv2 dataset, we directly apply the model trained on BSDS500 and report the performance on the 400 tests to evaluate the generality of the model. As for the two medical datasets, since their samples have considerable difference, therefore, we train and test the models on their respective data splits.

\vspace{0.1in}
\noindent{\textbf{Augmentation via Patch Jitter.}}
To further improve the performance and enhance the generality of our model, we propose to augment the data by jittering the image patches. Specifically, the proposed patch jitter augmentation comprises two components, \ie, patch shuffle and random shift. Fig.~\ref{vis_dataAug} gives one example for each type of augmentation.
%Fig.~\ref{vis_dataAug} shows the respective examples for these two types of data augmentation. The \textbf{patch shuffle} first samples two image patches with shape $S\times S$ and then randomly exchange them to extend the image patterns, the corresponding ground-truth patches are also exchanged accordingly to maintain the consistency. To further augment the data, we randomly pick up one 
%of the selected two patches and replace it with a random patch, whose ground-truth is assigned with a new label. While the \textbf{random shift} could be conducted along with the horizontal or vertical directions. For horizontal random shift, we first randomly sample a patch with shape $S\times L$, where $L=\text{rand\_int}(S, W)$, and a random offset $o=\text{rand\_int}(0,S)$. Then, we conduct a cycle translation on the patch by $o$ offset towards left or right. Meanwhile, the random patch trick in patch shuffle could also be adopted. Finally, the augmentation is done by replacing the original patch with the new one. Analogously, the augmentation along vertical direction could be done similarly. The patch jitter augmentation is repeated 2 times during training. SSN~\cite{SCN} is adopted as our baseline method in our experiments.

\noindent{\textbf{Patch Shuffle}} Assume $S$ is the superpixel sampling interval, for an image, we randomly pick up two patches $\{U,V\}$ with shape $S\times S$ and exchange their positions. The corresponding label patches are also exchanged accordingly. After that,  we randomly select one cell from $\{U,V\}$, and 1) keep it unchanged with probability 0.75  or 2) replace it using a random patch from normal distribution with probability 0.25. To maintain the consistency between image and label, we expand the label dimension and assign the random patch with the new label.

\noindent{\textbf{Random Shift}} The proposed random shift augmentation could be applied in two directions, \ie, horizontal and vertical. For convenience sake, we take the horizontal random shift as an example to elaborate the process. we first random sample a cell $C$ with shape $S\times L$, where $L=\text{rand\_int}(S, W)$, then the sampled cell $C$ is translated by a random offset $o=\text{rand\_int}(0,S)$. The translation could be conducted along two directions, \ie, the left and the right. Consequently, two new patches could be produced: $C_{l} = [C_{:,o:},C_{:,:o}]$ for left translation and $C_{r} = [C_{:,-o:},C_{:,:-o}]$ for right translation, where $[\cdot,\cdot]$ means concatenating along width. To further extend the image pattern, we also employ the same random patch trick in patch shuffle, that is, the $C_{:,:o}$ or the $C_{:,-o:}$ could be replaced by a random patch with probability 0.25. Finally, the augmentation is done by random replacing the $C$ with $C_l$ or $C_r$. Analogously, the augmentation along vertical direction could be done similarly. In our experiments, horizontal and vertical augmentation is randomly conducted. 

The patch jitter augmentation is repeated 2 times during training, and SCN~\cite{SCN} is adopted as our baseline method in our experiments.

\begin{figure*}[t]
\setlength{\abovecaptionskip}{-1pt} 
\setlength{\belowcaptionskip}{-7pt} 
\begin{center}
    \subfigure[\scriptsize BR-BP on ISIC-2017]{
    \begin{minipage}[t]{0.23\linewidth}
        \includegraphics[width=1.67in]{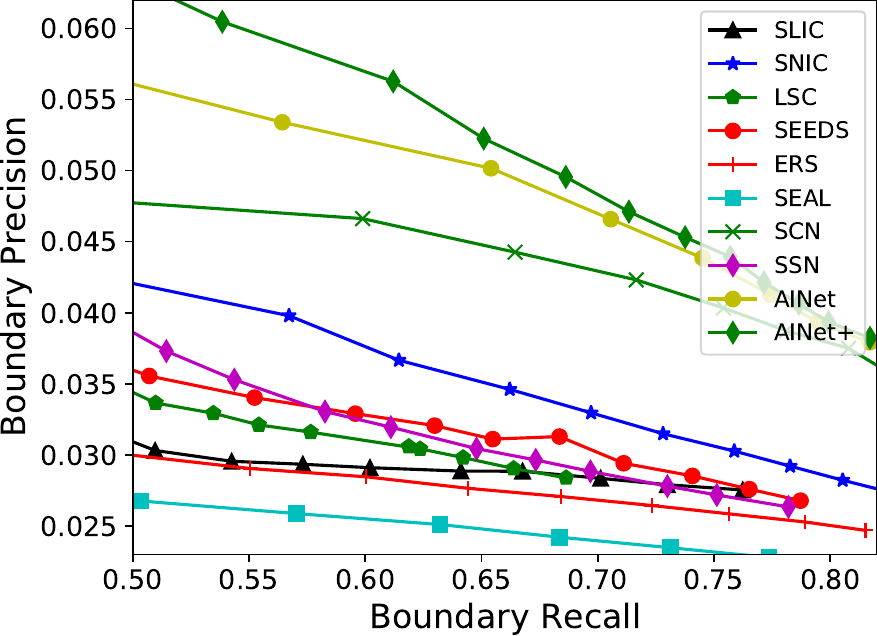}\\
    \vspace{-0.2cm}
    \end{minipage}
    }
    \hspace{0.02cm}
    \subfigure[\scriptsize ASA on ISIC-2017]{\begin{minipage}[t]{0.23\linewidth}
    \includegraphics[width=1.71in,height=1.2in]{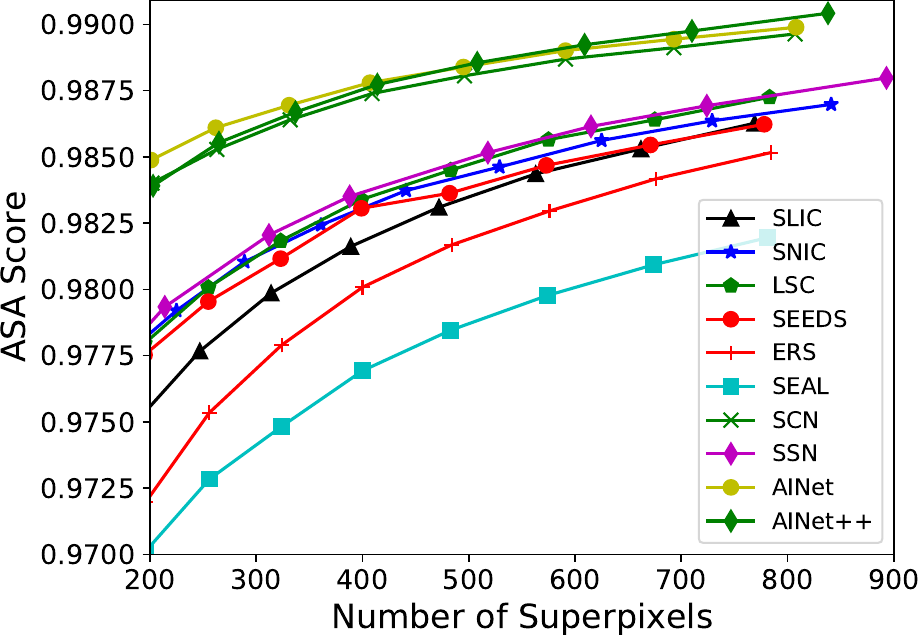}\\
    %\includegraphics[width=1.30in]{image_edit_outpainting/original_sketch_competing/yzx_528.jpg}\\
    %\vspace{0.01cm}
    %\includegraphics[width=1.30in]{image_edit_outpainting/original_sketch_competing/yzx_873.jpg}
    %\vspace{0.01cm}
    \vspace{-0.2cm}
    \end{minipage}}
    \hspace{0.02cm}
    \subfigure[\scriptsize BR-BP on ACDC]{
    \begin{minipage}[t]{0.23\linewidth}
        \includegraphics[width=1.67in,height=1.20in]{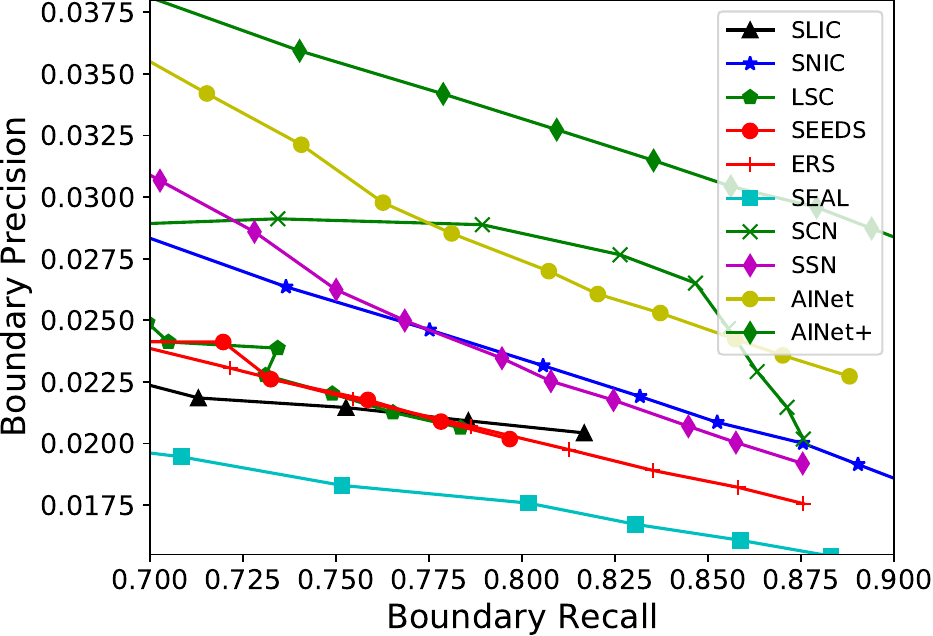}\\
    \vspace{-0.2cm}
    \end{minipage}
    }
    %\hspace{0.01cm}
    \subfigure[\scriptsize ASA on ACDC]{
    \begin{minipage}[t]{0.23\linewidth}
    \includegraphics[width=1.67in,height=1.2in]{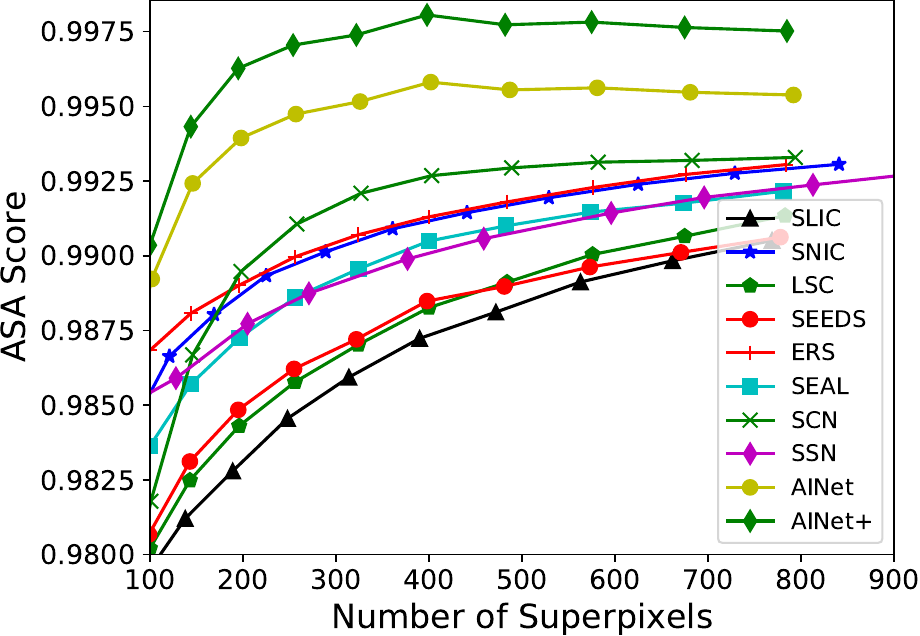}\\
    %\includegraphics[width=1.30in]{image_edit_outpainting/original_sketch_competing/our_528.jpg}\\
    %\vspace{0.02cm}
    %\includegraphics[width=1.30in]{image_edit_outpainting/original_sketch_competing/our_873.jpg}
    \vspace{-0.2cm}
    \end{minipage}
    }
\end{center}
\vspace{-0.3cm}
 \caption{Performance comparison on datasets ISIC-2017 and ACDC.}
\label{main_perform_medical}
\end{figure*}

\begin{figure*}[t]
\setlength{\belowcaptionskip}{-4pt} 
\begin{center}
    \subfigure[Inputs]{
    \begin{minipage}[t]{0.14\linewidth}
    \includegraphics[width=1.12in,height=1.36in]{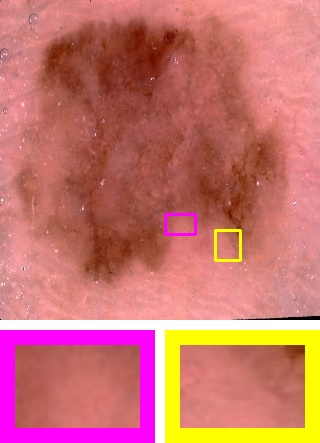}\\
    \vspace{-0.3cm}
    \includegraphics[width=1.12in,height=1.36in]{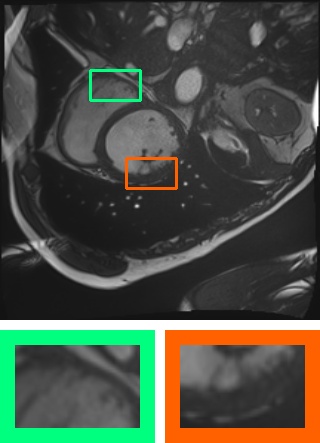}\\
     %\vspace{-0.4cm}
    %\includegraphics[width=1.35in]{tcyb/original_sketch_competing/pix2pix_528.jpg}\\
    \vspace{-0.3cm}
    \end{minipage}
    }
   \hspace{0.02cm}
    \subfigure[GT label]{
    \begin{minipage}[t]{0.15\linewidth}
    \includegraphics[width=1.12in,height=1.36in]{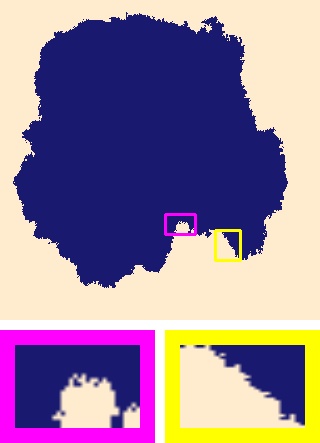}\\
     \vspace{-0.3cm}
    %\includegraphics[width=1.35in]{tcyb/original_sketch_competing/yzx_873.jpg}\\
     %\vspace{-0.4cm}
    \includegraphics[width=1.12in,height=1.36in]{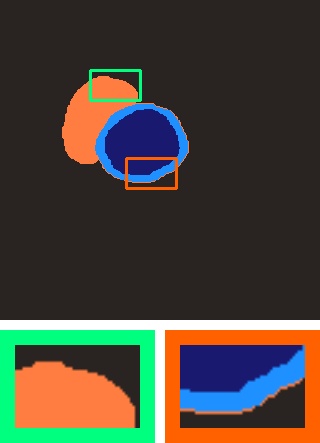}\\
     \vspace{-0.3cm}
    \end{minipage}}
    \hspace{-0.06cm}
    \subfigure[SCN~\cite{SCN}]{
    \begin{minipage}[t]{0.15\linewidth}
    \includegraphics[width=1.12in,height=1.36in]{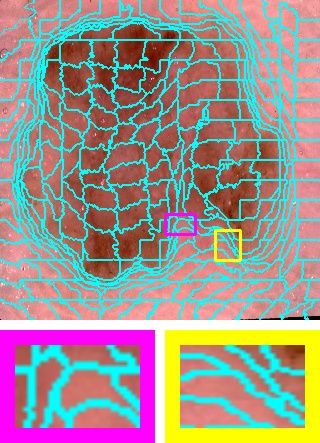}\\
     \vspace{-0.3cm}
    %\includegraphics[width=1.35in]{tcyb/original_sketch_competing/BDIE_873.png}\\
    % \vspace{-0.4cm}
    \includegraphics[width=1.12in,height=1.36in]{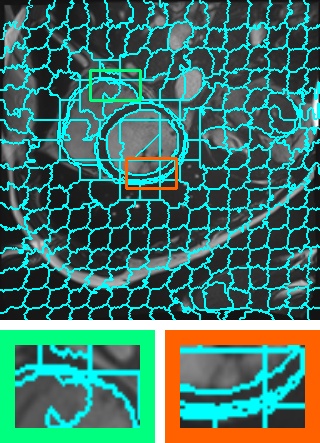}\\
     \vspace{-0.3cm}
    \end{minipage}
    }
    \hspace{-0.18cm}
    \subfigure[SSN~\cite{SSN}]{
    \begin{minipage}[t]{0.15\linewidth}
    \includegraphics[width=1.12in,height=1.36in]{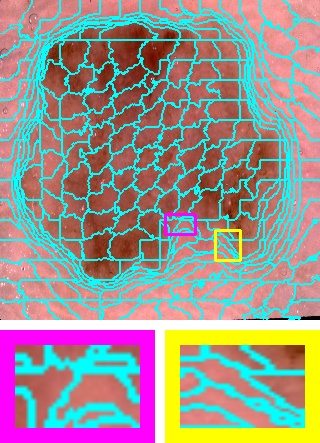}\\
     \vspace{-0.3cm}
    %\includegraphics[width=1.35in]{tcyb/original_sketch_competing/our_873.jpg}\\
    % \vspace{-0.4cm}
    \includegraphics[width=1.12in,height=1.36in]{medical_superpixel/ACDC/SSN_ACDC-patient014_frame13_2.jpg}\\
     \vspace{-0.3cm}
    \end{minipage}
    }
    \hspace{-0.18cm}
    \subfigure[AINet]{
    \begin{minipage}[t]{0.15\linewidth}
    \includegraphics[width=1.12in,height=1.36in]{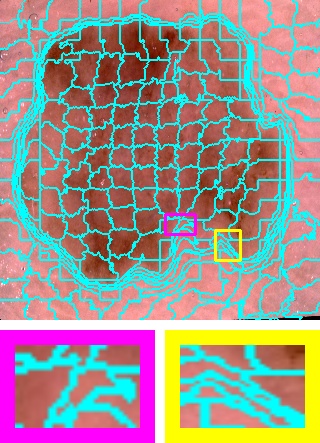}\\
    \vspace{-0.3cm}
    %\includegraphics[width=1.35in]{tcyb/original_sketch_competing/ori_873.jpg}\\
    % \vspace{-0.4cm}
    \includegraphics[width=1.12in,height=1.36in]{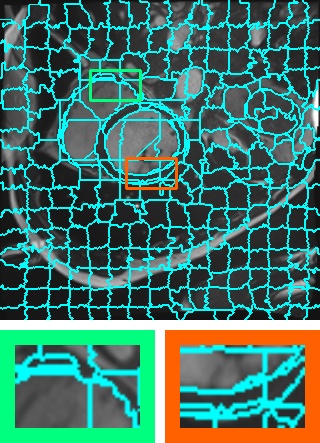}\\
     \vspace{-0.3cm}
    \end{minipage}
    }
     \hspace{-0.18cm}
    \subfigure[AINet+]{
    \begin{minipage}[t]{0.15\linewidth}
    \includegraphics[width=1.12in,height=1.36in]{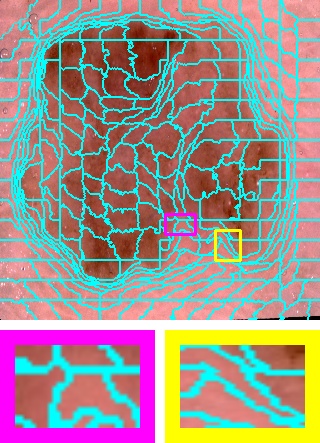}\\
    \vspace{-0.3cm}
    %\includegraphics[width=1.35in]{tcyb/original_sketch_competing/ori_873.jpg}\\
    % \vspace{-0.4cm}
    \includegraphics[width=1.12in,height=1.36in]{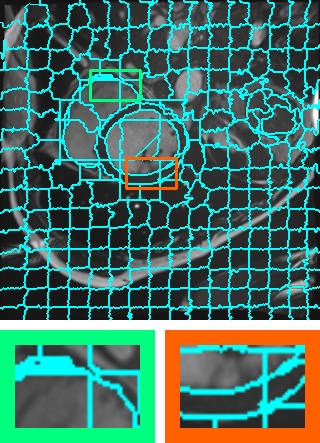}\\
     \vspace{-0.3cm}
    \end{minipage}
    }
\end{center}
\vspace{-0.6cm}
 \caption{Qualitative results of four SOTA superpixel methods, SCN, SSN, AINet~\cite{ainet} and our AINet+. The top row exhibits the results from ISIC-2017 dataset, while the bottom row shows the superpixels on ACDC dataset.}
 %\vspace{-0.2cm}
\label{main_visual_medical}
\end{figure*}

\begin{figure}[t]
\setlength{\abovecaptionskip}{-4pt} 
\setlength{\belowcaptionskip}{-4pt} 
\begin{center}
    \subfigure{
    \begin{minipage}[t]{0.43\linewidth}
        \includegraphics[width=1.61in]{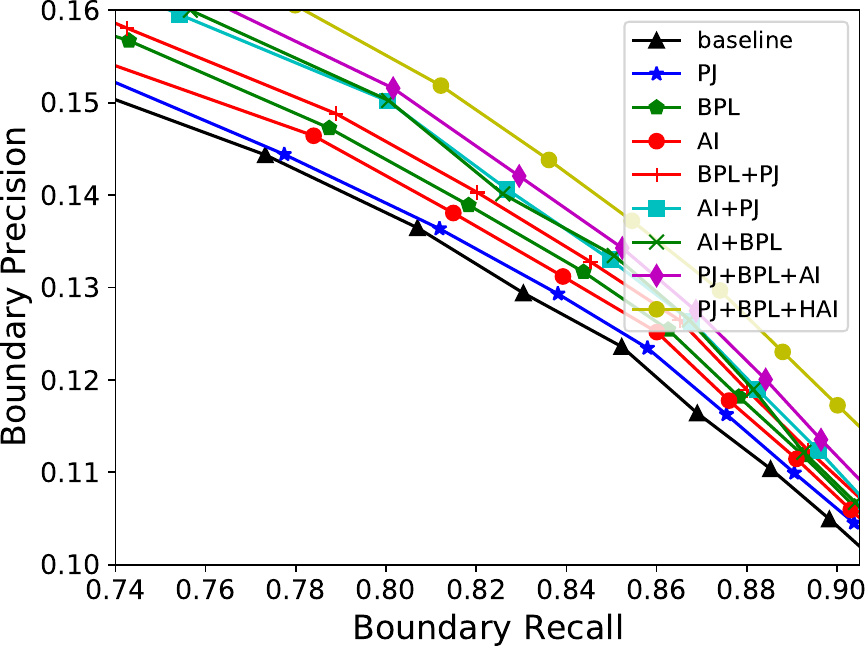}\\
    \vspace{-0.15cm}
    \end{minipage}
    }
    \hspace{0.01cm}
    \subfigure{\begin{minipage}[t]{0.43\linewidth}
    \includegraphics[width=1.61in]{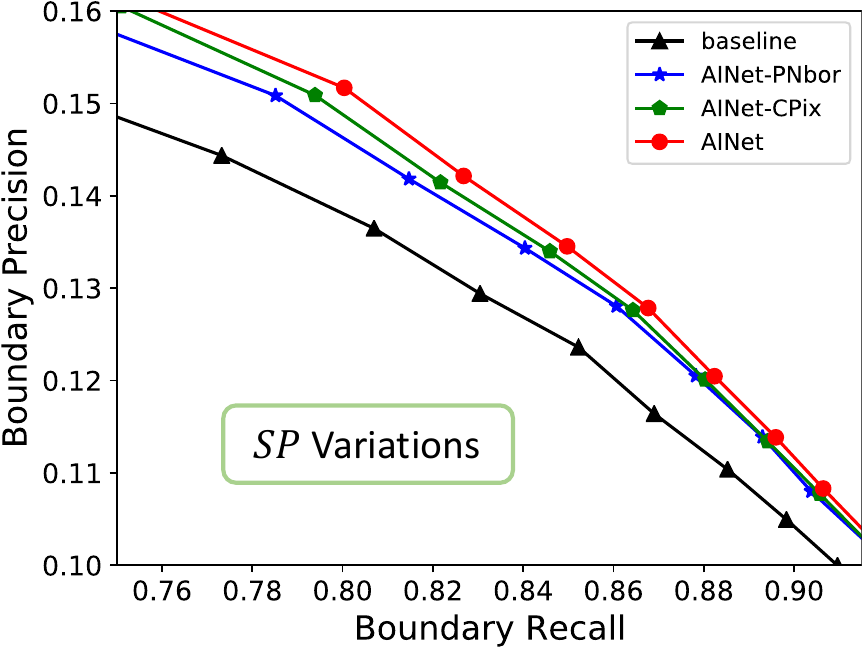}\\
    %\vspace{0.01cm}
    %\includegraphics[width=1.30in]{image_edit_outpainting/original_sketch_competing/yzx_873.jpg}
    %\vspace{0.01cm}
    \vspace{-0.15cm}
    \end{minipage}
    }
\end{center}
\vspace{-0.4cm}
 \caption{Ablation study on BSDS500, where the AI, BPL, PJ and HAI refer to the association implantation, boundary perceiving loss, patch jitter and the hierarchical association implantation module. The left figure shows the contributions of each component in our system, while the right one discusses two variations of $SP$ (Eq.~\ref{expanded_pixel}).}
\label{ablation_study}
\end{figure}

\begin{figure*}[t]
\begin{center}
\subfigure{
    \begin{minipage}[t]{0.15\linewidth}
    \includegraphics[width=1.12in]{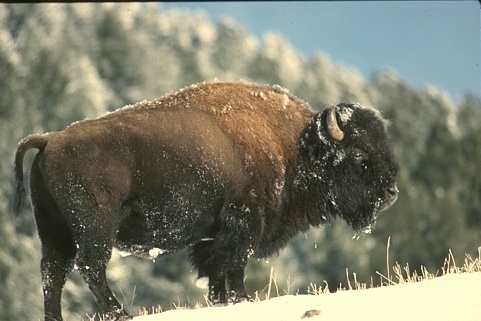}\\
    \vspace{-0.36cm}
    \includegraphics[width=1.12in]{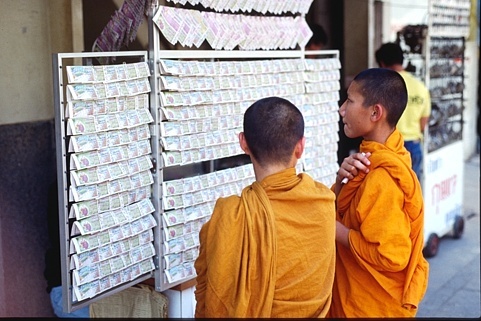}\\
    \vspace{-0.36cm}
    \includegraphics[width=1.12in]{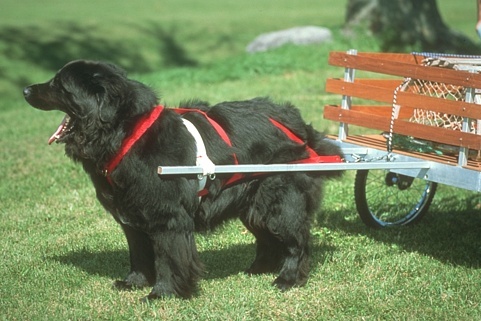}\\
     \vspace{-0.65cm}
     \mycaption{(a) Images}
    \end{minipage}
    }
   \hspace{-0.18cm}
    \subfigure{
    \begin{minipage}[t]{0.15\linewidth}
    \includegraphics[width=1.12in]{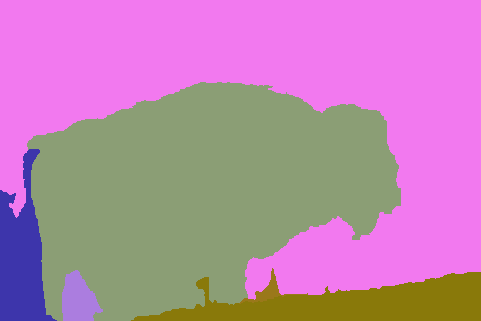}\\
     \vspace{-0.36cm}
     \includegraphics[width=1.12in]{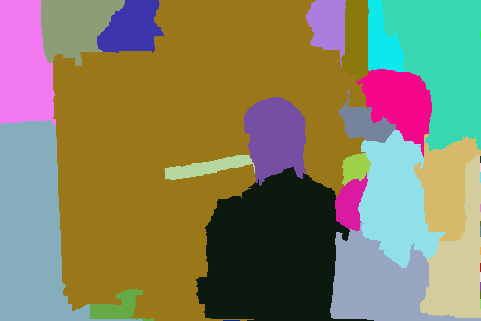}\\
     \vspace{-0.36cm}
    %\includegraphics[width=1.35in]{tcyb/original_sketch_competing/yzx_873.jpg}\\
     %\vspace{-0.4cm}
    \includegraphics[width=1.12in]{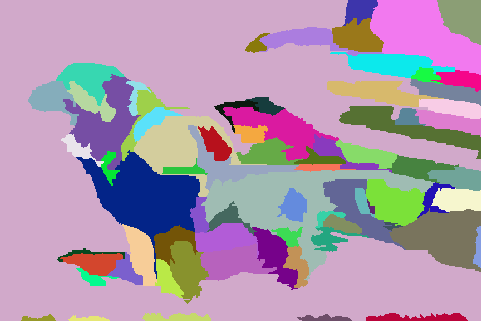}\\
     \vspace{-0.65cm}
     \mycaption{(c) SCN~\cite{SCN}}
    \end{minipage}}
    \hspace{-0.06cm}
    \subfigure{
    \begin{minipage}[t]{0.15\linewidth}
    \includegraphics[width=1.12in]{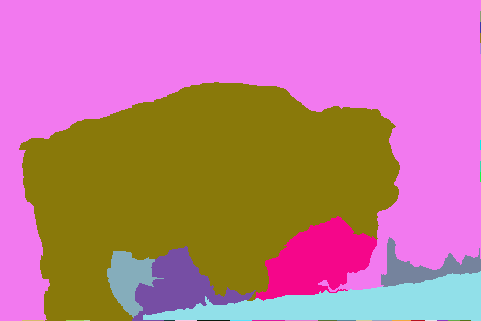}\\
     \vspace{-0.36cm}
    \includegraphics[width=1.12in]{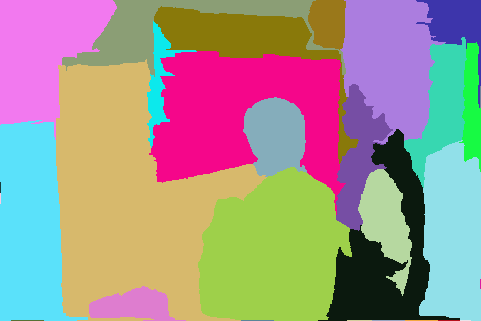}\\
     \vspace{-0.36cm}
    \includegraphics[width=1.12in]{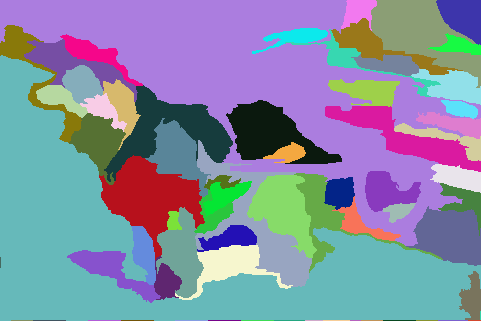}\\
     \vspace{-0.65cm}
     \mycaption{(d) SSN~\cite{SSN}}
    \end{minipage}
    }
    \hspace{-0.18cm}
    \subfigure{
    \begin{minipage}[t]{0.15\linewidth}
    \includegraphics[width=1.12in]{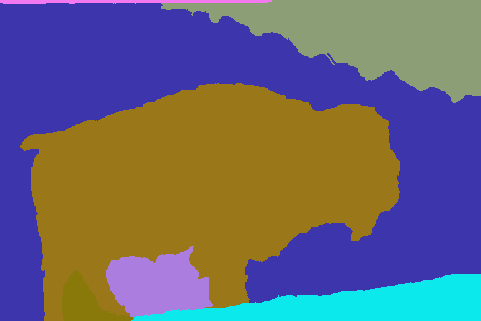}\\
     \vspace{-0.36cm}
    \includegraphics[width=1.12in]{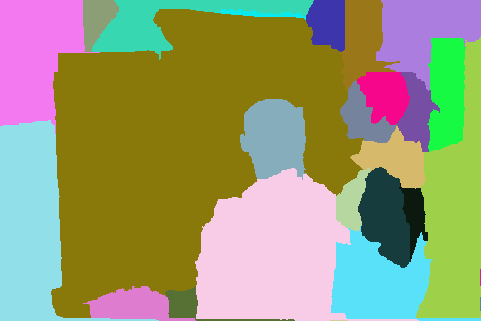}\\
    \vspace{-0.36cm}
    \includegraphics[width=1.12in]{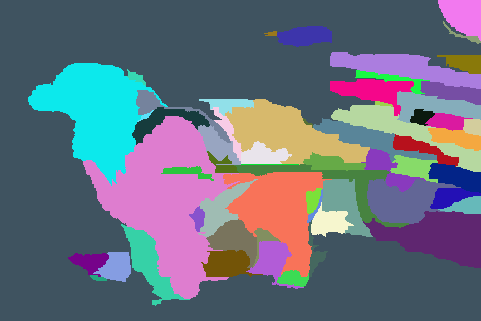}\\
     \vspace{-0.65cm}
     \mycaption{(e) AINet}
    \end{minipage}
    }
     \hspace{-0.18cm}
    \subfigure{
    \begin{minipage}[t]{0.14\linewidth}
    \includegraphics[width=1.12in]{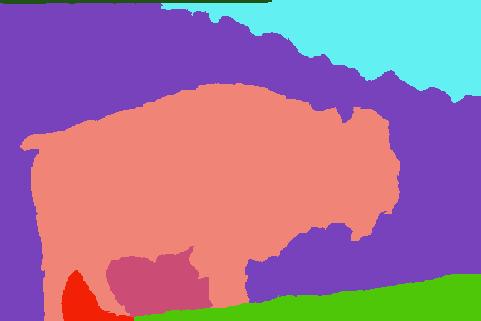}\\
    \vspace{-0.36cm}
    \includegraphics[width=1.12in]{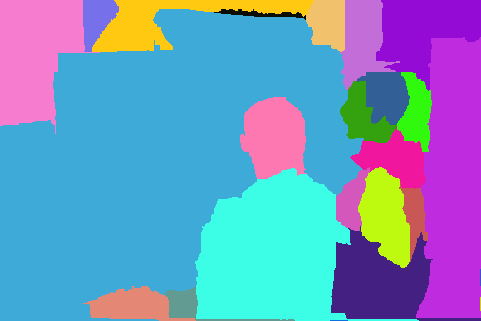}\\
    \vspace{-0.36cm}
    \includegraphics[width=1.12in]{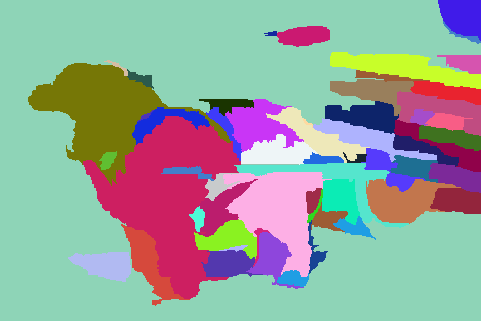}\\
     %\vspace{-0.4cm}
    %\includegraphics[width=1.35in]{tcyb/original_sketch_competing/pix2pix_528.jpg}\\
    \vspace{-0.65cm}
    \mycaption{(b) AINet+}
    \end{minipage}
    }
    \hspace{0.02cm}
    \subfigure{
    \begin{minipage}[t]{0.15\linewidth}
    \includegraphics[width=1.12in]{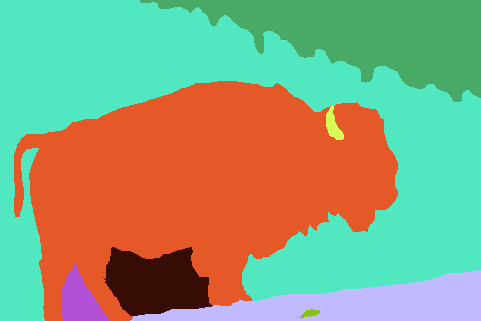}\\
    \vspace{-0.36cm}
   \includegraphics[width=1.12in]{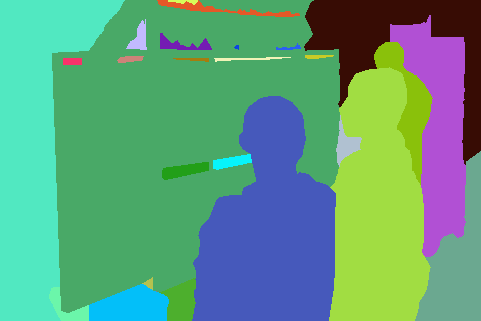}\\
    \vspace{-0.36cm}
    \includegraphics[width=1.12in]{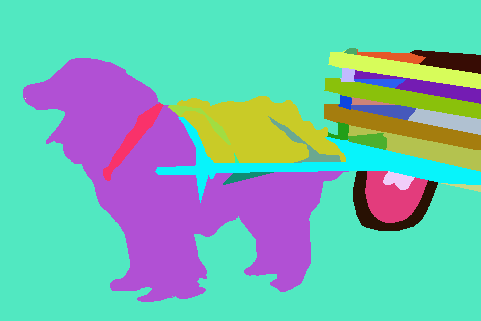}\\
     \vspace{-0.65cm}
     \mycaption{(f) GT label}
    \end{minipage}
    }
    \vspace{-0.03cm}
   \\ \centering\text{\normalsize I: The generated proposals from DEL~\cite{DEL} using different superpixels.} \\
   \vspace{0.1cm}

   \subfigure{
    \begin{minipage}[t]{0.15\linewidth}
    \includegraphics[width=1.12in]{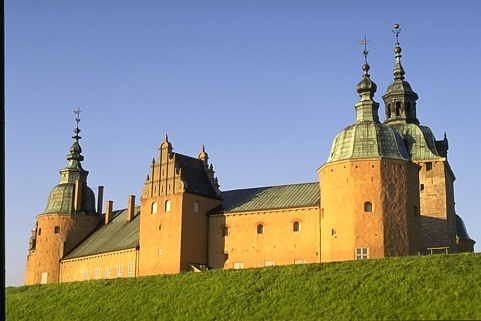}\\
    \vspace{-0.36cm}
    \includegraphics[width=1.12in]{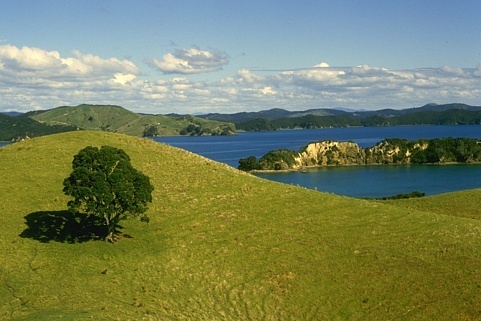}\\
    \vspace{-0.36cm}
    \includegraphics[width=1.12in]{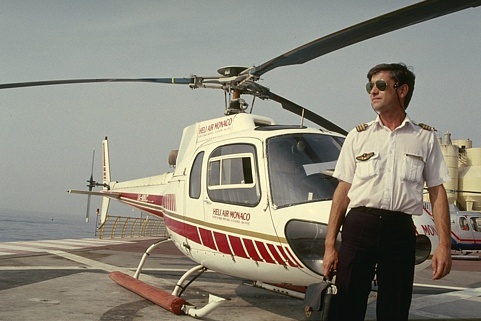}\\
     \vspace{-0.65cm}
     \mycaption{(a) Images}
    \end{minipage}
    }
    \hspace{-0.18cm}
    \subfigure{
    \begin{minipage}[t]{0.14\linewidth}
    \includegraphics[width=1.12in]{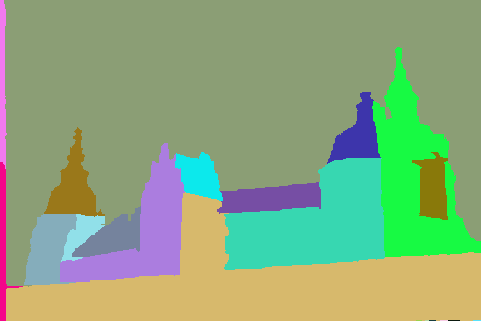}\\
    \vspace{-0.36cm}
    \includegraphics[width=1.12in]{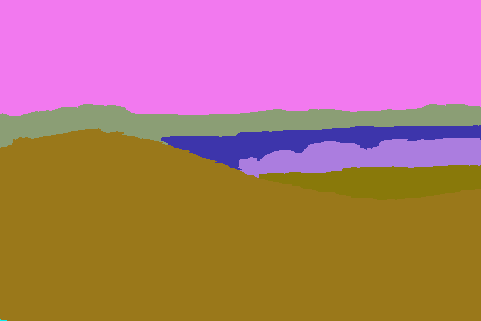}\\
    \vspace{-0.36cm}
    \includegraphics[width=1.12in]{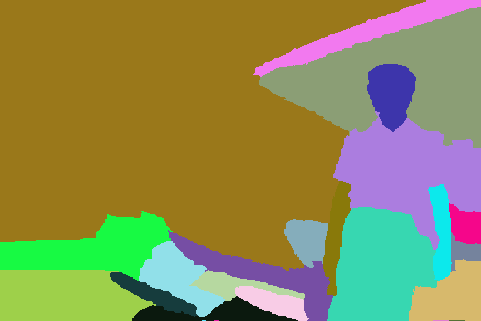}\\
     %\vspace{-0.4cm}
    %\includegraphics[width=1.35in]{tcyb/original_sketch_competing/pix2pix_528.jpg}\\
    \vspace{-0.65cm}
    \mycaption{(b) Threshold=0.3}
    \end{minipage}
    }
   \hspace{-0.005cm}
    \subfigure{
    \begin{minipage}[t]{0.15\linewidth}
    \includegraphics[width=1.12in]{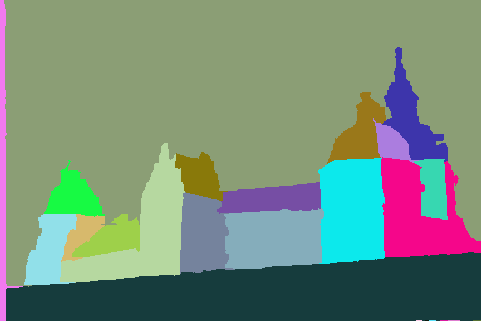}\\
     \vspace{-0.36cm}
     \includegraphics[width=1.12in]{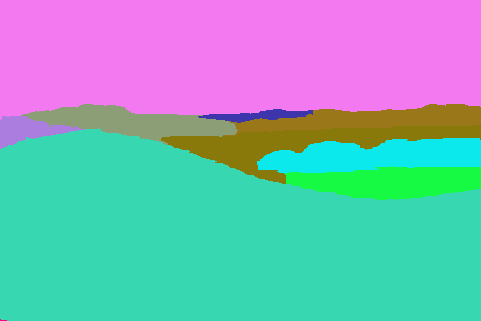}\\
     \vspace{-0.36cm}
    %\includegraphics[width=1.35in]{tcyb/original_sketch_competing/yzx_873.jpg}\\
     %\vspace{-0.4cm}
    \includegraphics[width=1.12in]{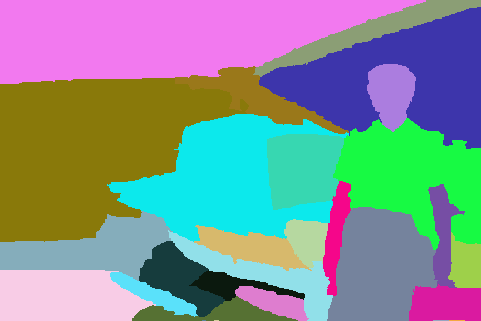}\\
     \vspace{-0.65cm}
     \mycaption{(c) Threshold=0.4}
    \end{minipage}}
    \hspace{-0.06cm}
    \subfigure{
    \begin{minipage}[t]{0.15\linewidth}
    \includegraphics[width=1.12in]{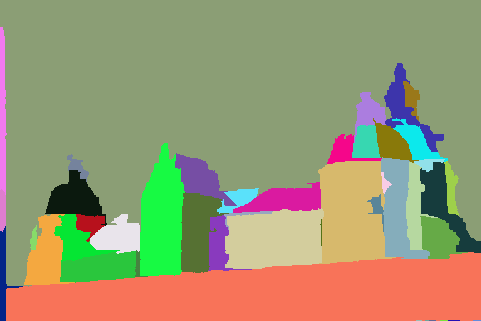}\\
     \vspace{-0.36cm}
    \includegraphics[width=1.12in]{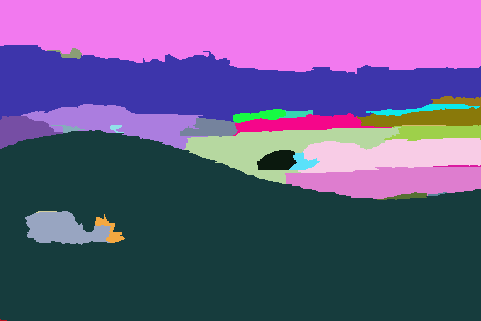}\\
     \vspace{-0.36cm}
    \includegraphics[width=1.12in]{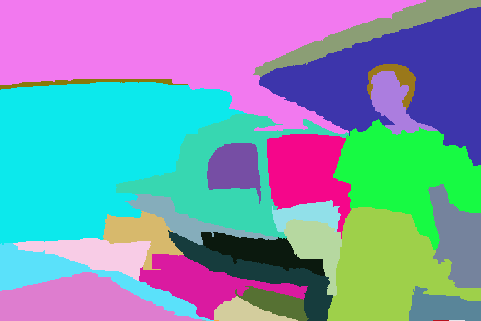}\\
     \vspace{-0.65cm}
     \mycaption{(d) Threshold=0.5}
    \end{minipage}
    }
    \hspace{-0.18cm}
    \subfigure{
    \begin{minipage}[t]{0.15\linewidth}
    \includegraphics[width=1.12in]{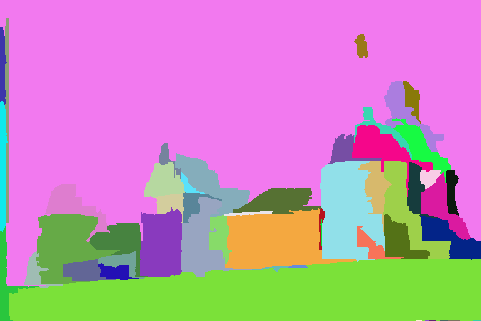}\\
     \vspace{-0.36cm}
    \includegraphics[width=1.12in]{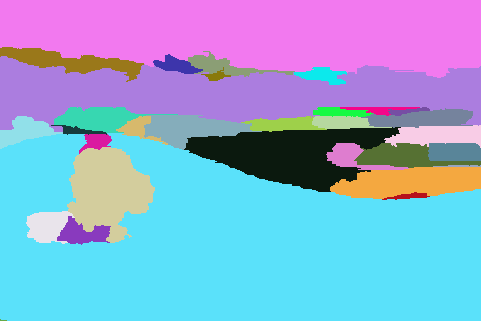}\\
     \vspace{-0.36cm}
    \includegraphics[width=1.12in]{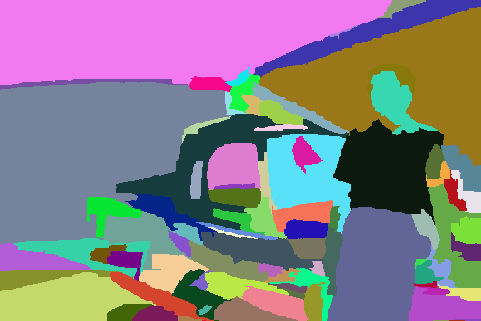}\\
     \vspace{-0.65cm}
     \mycaption{(e) Threshold=0.6}
    \end{minipage}
    }
    \hspace{-0.18cm}
    \subfigure{
    \begin{minipage}[t]{0.15\linewidth}
    \includegraphics[width=1.12in]{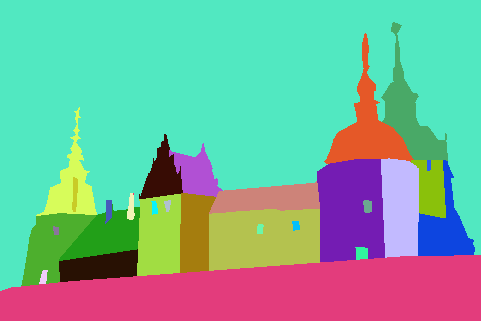}\\
    \vspace{-0.36cm}
   \includegraphics[width=1.12in]{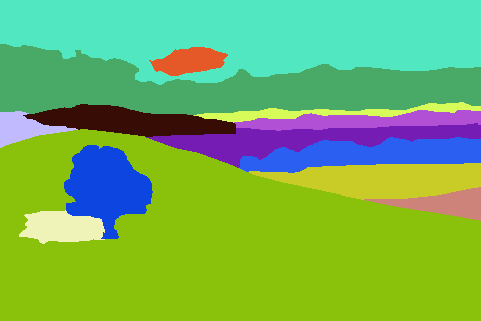}\\
    \vspace{-0.36cm}
    \includegraphics[width=1.12in]{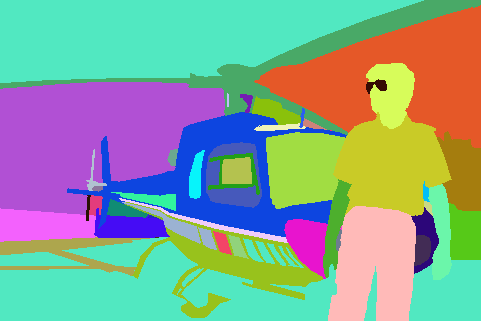}\\
     \vspace{-0.65cm}
     \mycaption{(f) GT label}
    \end{minipage}
    }
    \vspace{-0.03cm}
   \\ \centering\text{\normalsize II: The generated proposals using different thresholds (1 as upper bound).} \\
   \vspace{0.1cm}
\end{center}
\vspace{-0.5cm}
 \caption{Qualitative proposals from DEL~\cite{DEL} using different superpixels (I),  and the results of DEL~\cite{DEL} with our superpixel using different thresholds (II), where threshold=0.3 mean merging the adjacent superpixels if their similarity is above 0.3.}
 %\vspace{-0.2cm}
\label{del_dis}
\end{figure*}

\vspace{0.1in}
\noindent{\textbf{Implementation Details.}}
In training stage, the image is randomly cropped to 256$\times$256 as input, and the network is trained using the adam optimizer~\cite{adam} for 4k iterations with batch size 16. The learning rate starts with 8e-5 and is discounted by 0.5 for every 2K iterations. The sampling interval is fixed as 16, consequently, the encoder part employs 4 convolution\&pooling operations to get the superpixel embedding $M$ with shape $14\times 14\times 256$. The following decoder module produces the pixel embedding with shape $256\times 256\times 16$ using 4 convolution\&deconvolution operations. Then, the channels of superpixel embedding are first compressed by two convolution layers: 256$\Rightarrow$64$\Rightarrow$16, then our AI module is performed.
Our AI module could be effectively implemented by a repeat and mask operation. Specifically, each pixel embedding in $E$ is repeated to form a $3\times 3\times D$ cubic, thus, the expanded $E$ (marked as $E^{*}$) is with shape $(3H)\times (3W)\times D$.  Next, we slide a $3\times 3$ window on Sp embedding and repeat the overall $3\times 3$ patch by $16\times 16$ times, since the pixels in the same grid share the 9 neighbor grids (see Fig.~\ref{first_shown}), consequently, the shape of the expanded Sp embedding (marked as $M^{*}$) is also $(3H)\times (3W)\times D$. A mask $G$ is pre-defined by repeating $3\times 3$ matrix [0,0,0; 0,1,0; 0,0,0] $H\times W$ times. Then, the implantation operation (Eq.5) is implemented: $\hat{F}=G\odot E^{*}+M^{*}$. Finally, a $3\times 3$ conv with stride 3 is applied on $\hat{F}$ to get the output $E^{'}$, which is fed forward to the next layer.

To perform the association module on the second-last layer, we first down-sample the bottleneck feature $M$ by a $3\times 3$ convolution with stride 2, and adjust its channels by two convolution layers: 512$\Rightarrow$128$\Rightarrow$32, and the AI module is analogously performed.
The boundary-perceiving loss also acts on the pixel embedding, where the patch size is set to 5, \ie, $K=5$.  In the following, two convolution layers are stacked to predict the association map $Q$ with shape $256\times 256\times 9$. In our conference version, the input resolution is 208$\times$208, we enlarge the resolution as 256 in this work to enable the downsampling to appropriately go one more step, such that the association implantation could be performed on the second-last layer. But it is worthy noting that simply enlarging the resolution did not improve the performance in our experiments, therefore, the performance gain is not from the larger resolution.  
In our practice, simultaneously equipping the boundary-perceiving loss and AI Module could not make the performance step further, therefore, we first train the network using the first two items in Eq.~\ref{total_loss} for 3K iterations, and use the boundary-perceiving loss to finetune 1K. Following Yang's practice ~\cite{SCN}, the weight of position reconstruction loss is set to 0.003/16, while the weight for our boundary-perceiving loss is fixed to 0.5, \ie, $\lambda=0.003/16, \alpha=0.5$. In testing, we employ the same strategy as~\cite{SCN} to produce varying numbers of superpixels. For example, if we want to produce 400 superpixels, the image would be resized to $320\times 320$.

Several methods are considered for performance comparison, including classic methods, SLIC~\cite{SLIC}, LSC~\cite{LSC}, ERS~\cite{ERS}, SEEDS~\cite{SEEDS}, SNIC~\cite{SNIC} and deep learning-based methods, SEAL~\cite{SEAL}, SSN~\cite{SSN}, SCN~\cite{SCN}. We simply use the OpenCV implementation for methods SLIC, LSC and SEEDS. For other methods, we use the official implementations with the recommended parameters from the authors. AINet~\cite{ainet} is the conference version of our method, while AINet+ refers to AINet equipped with the hierarchical association implantation, \ie, the method proposed in this paper. 

\vspace{0.1in}
\noindent{\textbf{Evaluation Metrics.}} We use three popular metrics including achievable segmentation accuracy (ASA), boundary recall (BR) and boundary precision (BP) to evaluate the performance of superpixel. ASA score studies the upper bound on the achievable segmentation accuracy using superpixel as pre-processing step, while BR and BP focus on accessing how well the superpixel model could identify the semantic boundaries. The higher value of these metrics indicates better superpixel segmentation performance.
\subsection{Comparison with the state-of-the-art methods}
%Fig.~\ref{main_perform} exhibits the results on BSDS500 and NYUv2 test sets, 
Fig.~\ref{main_perform} reports the quantitative comparison results on BSDS500 and NYUv2 test sets. As indicated in Fig.~\ref{main_perform}, our AINet+ attains the best ASA score and BR-BP on both datasets. 
With the help of deep convolution networks, the methods, SEAL, SCN, SSN, AINet and AINet+ could achieve superior or comparable performance against the traditional superpixel algorithms, and our AINet and AINet+ are the top 2 models among them.  
From Fig.~\ref{main_perform} (a)-(b), the AINet could surpass the traditional methods by a large margin on BSDS500 dataset. By harvesting the pixel-sueprpixel level context and highlighting the boundaries, AINet could also outperform the deep methods SEAL, SCN and SSN. The hierarchical AI module further boosts the performance, our AINet+ achieves the most outstanding performance. Fig.~\ref{main_perform} (c)-(d) shows the performance when adapting to the NYUv2 test set, we can observe that the AINet and AINet+ also shows better generality.   Although the BR-BP is comparable with the SCN and SSN, ASA scores of AINet and AINet+ are more outstanding than all of the competitive methods.

To further evaluate our method, we also conduct experiments on two widely-used medical benchmarks, \ie, ISIC-2017~\cite{ISIC2017} and ACDC~\cite{ACDC},  Fig.~\ref{main_perform_medical} exhibits the performance comparison on these two benchmarks. From Fig.~\ref{main_perform_medical}, we can see that the superiority of our AINet and AINet+ are clearer, both AINet and AINet+ could surpass all competing methods by a large margin on both the BR-BP and ASA score. Comparing to AINet, our AINet+ could attain better results.

Fig.~\ref{main_visual} and Fig.~\ref{main_visual_medical} subsequently show the qualitative results of four state-of-the-art methods on dataset BSDS500, NYUv2, ISIC-2017 and ACDC,  comparing to the competitive methods, the boundaries of our results are more accurate and clearer, which intuitively shows the superiority of our method.

\begin{figure}[t]
\setlength{\abovecaptionskip}{-1pt} 
\setlength{\belowcaptionskip}{-3pt} 
\begin{center}
    \includegraphics[width=2.6in]{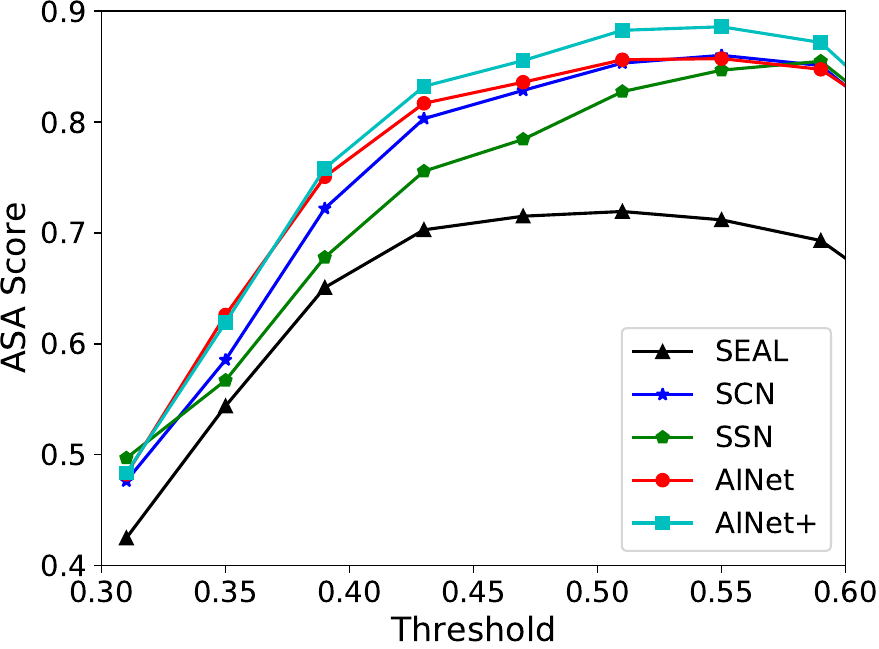}\\
    \vspace{-0.15cm}
\end{center}

 \caption{The ASA scores of four state-of-the-art methods on object proposal generation.}
\label{obp}
\end{figure}

\begin{figure}[t]
\setlength{\abovecaptionskip}{-1pt} 
\setlength{\belowcaptionskip}{-5pt} 
\begin{center}
    \includegraphics[width=2.6in]{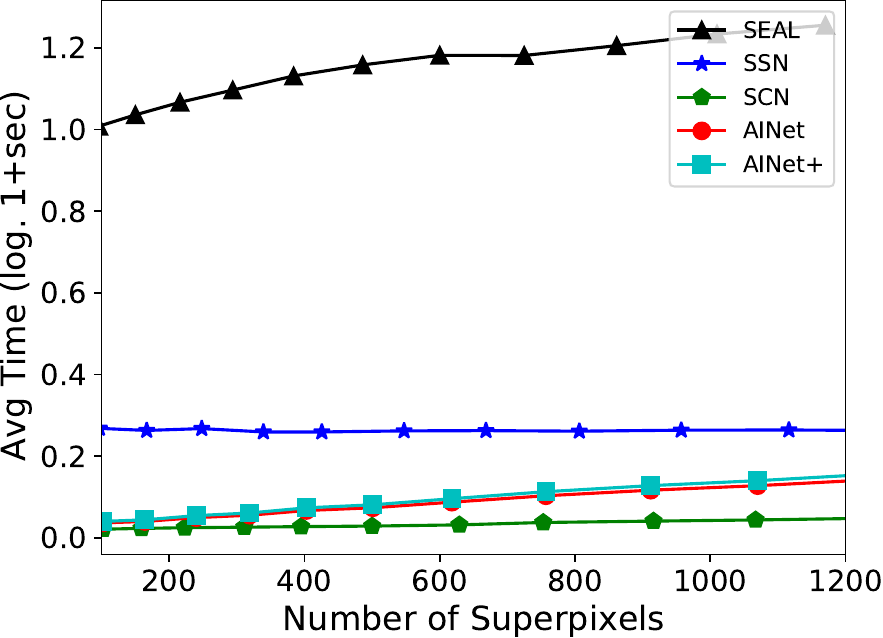}\\
    \vspace{-0.15cm}
\end{center}

 \caption{The average time costs of four deep learning based methods w.r.t number of superpixels. The runtime is added with 1 and scaled by logarithmic to show positive values and a clear tendency.}
\label{time_comparison}
\end{figure}
%\vspace{-20pt}

\begin{figure*}[t]
\setlength{\belowcaptionskip}{-4pt} 
\begin{center}
    \subfigure[Inputs]{
    \begin{minipage}[t]{0.18\linewidth}
    %\includegraphics[width=1.39in]{stereoMatch/sceneFlow/frames_cleanpass_A_0007_0008.jpg}\\
    %\vspace{-0.3cm}
    \includegraphics[width=1.39in]{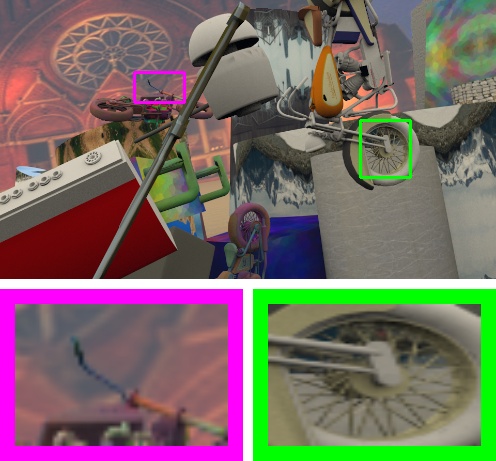}\\
    \vspace{-0.3cm}
    \includegraphics[width=1.39in]{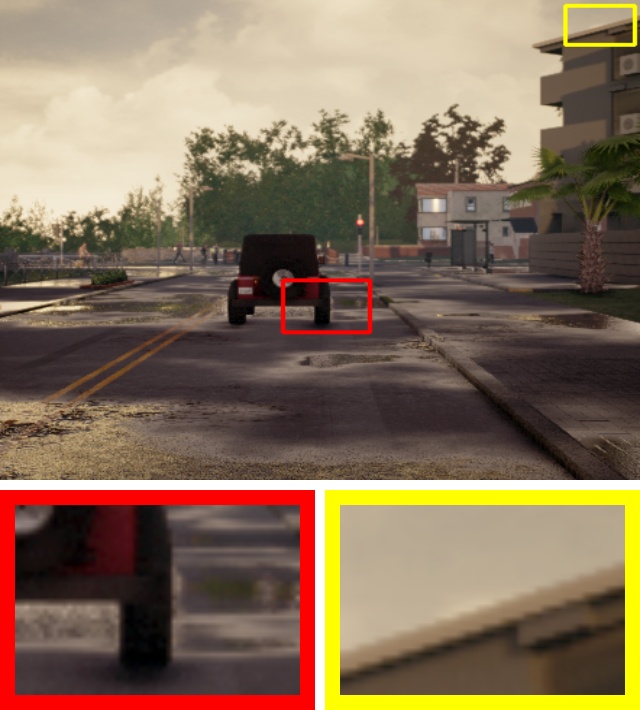}\\
    \vspace{-0.3cm}
    \end{minipage}
    }
   \hspace{-0.08cm}
    \subfigure[GT label]{
    \begin{minipage}[t]{0.18\linewidth}
    %\includegraphics[width=1.39in]{stereoMatch/sceneFlow/gt_frames_cleanpass_A_0007_0008.jpg}\\
     %\vspace{-0.3cm}
    %\includegraphics[width=1.35in]{tcyb/original_sketch_competing/yzx_873.jpg}\\
     %\vspace{-0.4cm}
    \includegraphics[width=1.39in]{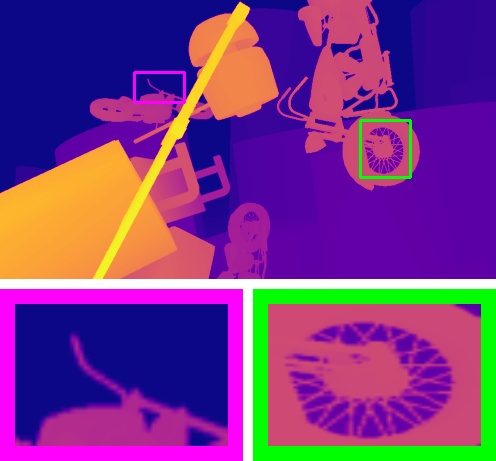}\\
     \vspace{-0.3cm}
     \includegraphics[width=1.39in]{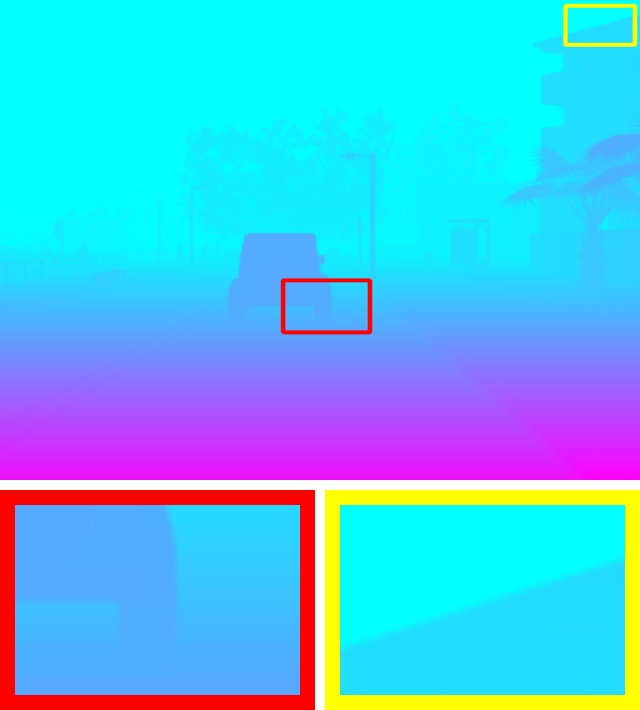}\\
     \vspace{-0.3cm}
    \end{minipage}}
    \hspace{-0.06cm}
    \subfigure[PSMNet]{
    \begin{minipage}[t]{0.18\linewidth}
    %\includegraphics[width=1.39in]{stereoMatch/sceneFlow/PSM_frames_cleanpass_A_0007_0008.jpg}\\
    % \vspace{-0.3cm}
    %\includegraphics[width=1.35in]{tcyb/original_sketch_competing/BDIE_873.png}\\
    % \vspace{-0.4cm}
    \includegraphics[width=1.39in]{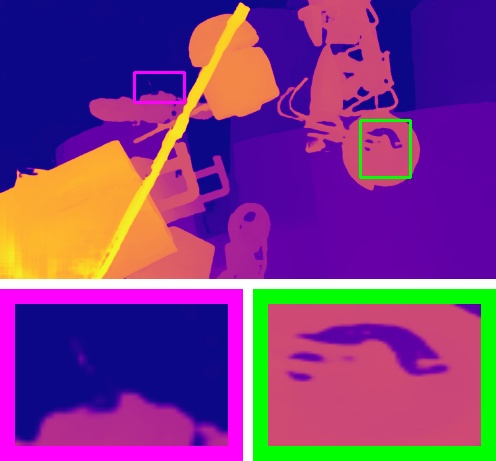}\\
     \vspace{-0.3cm}
     \includegraphics[width=1.39in]{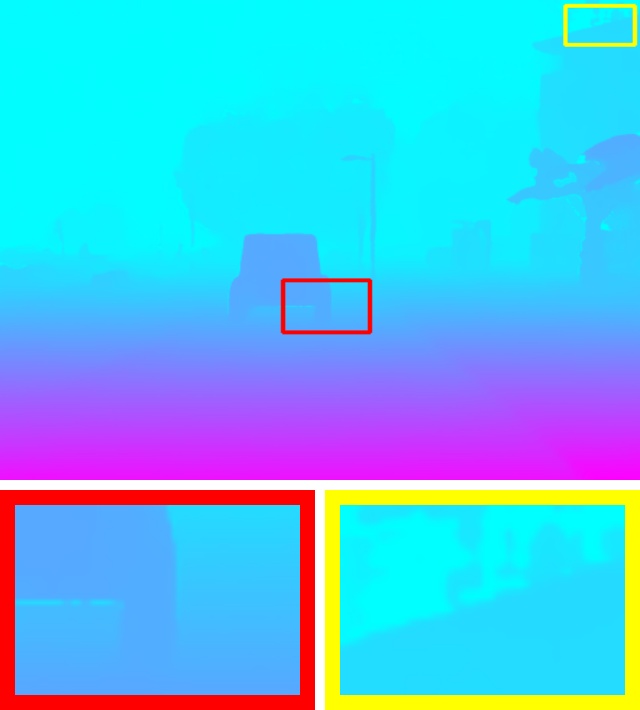}\\
     \vspace{-0.3cm}
    \end{minipage}
    }
    \hspace{-0.18cm}
    \subfigure[$\text{PSMNet}_{\text{SCN}}$]{
    \begin{minipage}[t]{0.18\linewidth}
    %\includegraphics[width=1.39in]{stereoMatch/sceneFlow/AINet_frames_cleanpass_A_0007_0008.jpg}\\
    % \vspace{-0.3cm}
    %\includegraphics[width=1.35in]{tcyb/original_sketch_competing/our_873.jpg}\\
    % \vspace{-0.4cm}
    \includegraphics[width=1.39in]{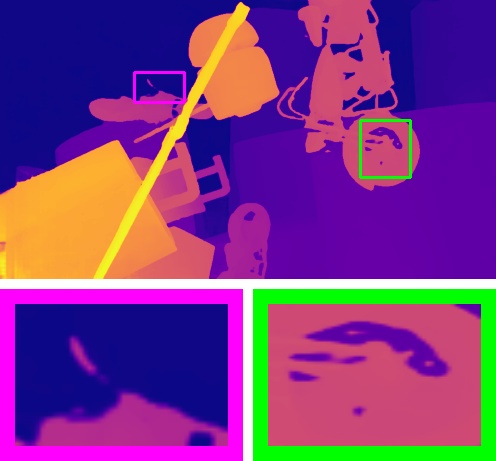}\\
     \vspace{-0.3cm}
     \includegraphics[width=1.39in]{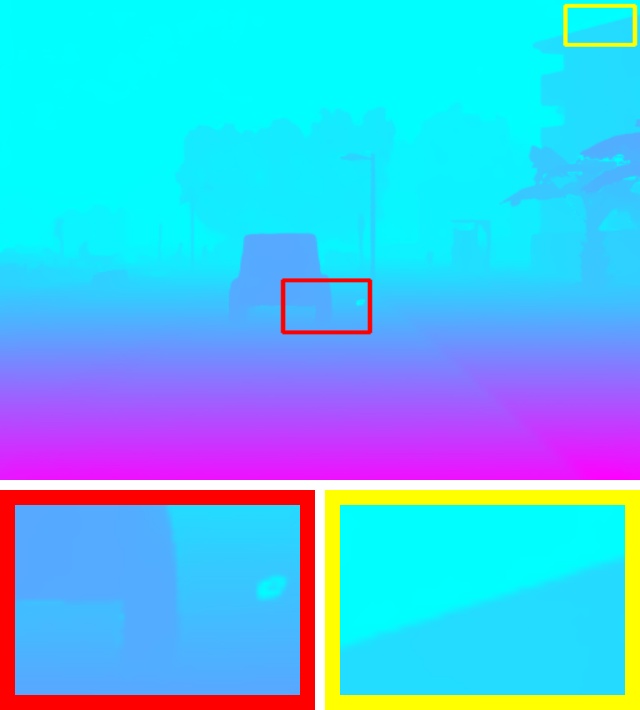}\\
     \vspace{-0.3cm}
    \end{minipage}
    }
    \hspace{-0.18cm}
    \subfigure[$\text{PSMNet}_{\text{AINet+}}$]{
    \begin{minipage}[t]{0.18\linewidth}
    %\includegraphics[width=1.39in]{stereoMatch/sceneFlow/Ours_frames_cleanpass_A_0007_0008.jpg}\\
    %\vspace{-0.3cm}
    %\includegraphics[width=1.35in]{tcyb/original_sketch_competing/ori_873.jpg}\\
    % \vspace{-0.4cm}
    \includegraphics[width=1.39in]{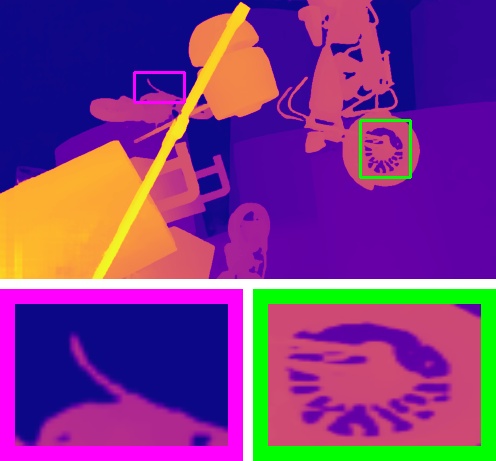}\\
     \vspace{-0.3cm}
     \includegraphics[width=1.39in]{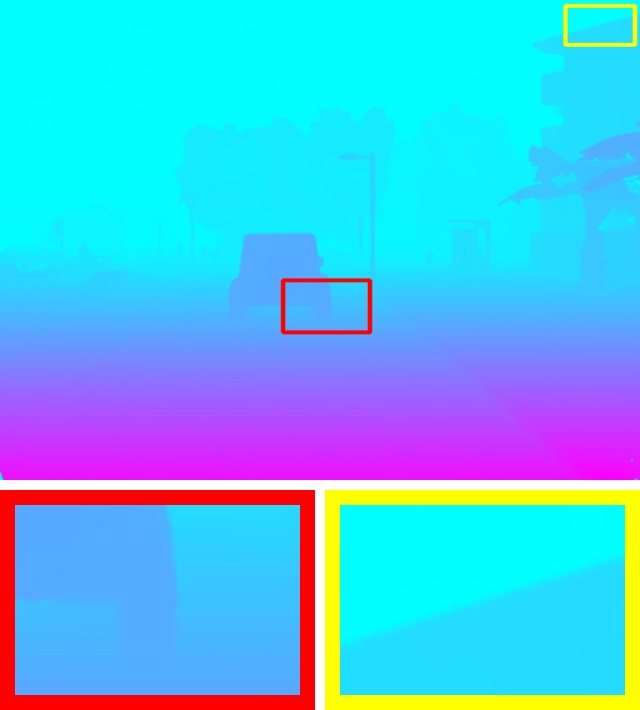}\\
     \vspace{-0.3cm}
    \end{minipage}
    }
\end{center}
\vspace{-0.6cm}
 \caption{Qualitative disparity predictions of three methods. The top row exhibits the results from SceneFlow dataset, while the bottom row shows the prediction on HR-VS dataset. By integrating the superpixels into the baseline PSM model, the results could be improved, and the PSM with our AINet+ outperforms the other approaches.}
 %\vspace{-0.2cm}
\label{main_visual_disparity}
\end{figure*}
\subsection{Ablation Study}
To validate the respective contributions of our proposed modules including the data augmentation trick, AI module, and the boundary-perceiving loss, we conduct ablation study on BSDS500 dataset to thoroughly study their effectiveness.  The left figure in Fig.~\ref{ablation_study} reports the performances of all methods, where the BPL means the boundary-perceiving loss, and BPL+PJ stands for the baseline simultaneously equipped with the boundary-perceiving loss and the patch jitter augmentation. From Fig.~\ref{ablation_study}, we can observe that individually applying the three modules on the baseline method could all boost the performance, and the boundary-perceiving loss could contribute the most performance gains. The combination of the patch jitter augmentation and the BPL or AI Module could make the performance step further, and the AI module equipped with the data augmentation achieves better performance. When simultaneously employing the three modules,  the performance gets further improved. By equipping the hierarchical AI operation,  we could harvest the best BR-BP. 
%The margins of ASA scores shown in Fig.~\ref{ablation_study} (b) are not obvious, since the value is high enough and hard to observe a significant improvement, even so, we could still observe that the model with all modules equipped achieve better ASA scores than the baseline method. 

Besides, we also give a discussion for two alternative choices of $SP$ (Eq.~\ref{expanded_pixel}): a greedy version of $SP$ that further adds the neighbor pixels to the corresponding surrounding superpixels like the central position, for example, $\hat{m}_{t}$ is replaced by $\hat{m}_{t}+e_{t}$; And a simplified version that ignoring the central superpixel, \ie, $\hat{m}_c+e_p$ changes to $e_p$. The models with the above two versions of $SP$ are marked as AINet-PNbor and AINet-CPix, respectively.  The right figure of Fig.~\ref{ablation_study} shows the results, we can observe that AINet-PNbor and AINet-CPix could both surpass the baseline but perform a litter worse than AINet. By summing the neighbor pixels, the AINet-PNbor could integrate the pixel-wise relation, on the other hand, the sum operation would also reduce the force of superpixel embedding, which would conspire against capturing the pixel-superpixel context. For AINet-CPix, the excluded  $\hat{m}_c$ is also one of the neighbor superpixels, directly abandoning $\hat{m}_c$ would fail to explicitly perceive the relation between pixel $e_p$ and central superpixel $\hat{m}_c$.  Consequently, the above two variations of $SP$ are both not effective to capture the super context.

\subsection{Inference Efficiency}
Besides the performance, the inference speed is also a concerned aspect. Therefore, we conduct experiments on BSDS500 dataset to investigate the inference efficiency of five deep learning-based methods. 
To make a fair comparison, we only count the time of network inference and post-processing steps (if available). All methods run on the same workstation with NVIDIA 1080Ti GPU and Intel E5 CPU.

The time costs of five deep learning-based methods, SEAL, SCN, SSN AINet and AINet+ are reported in Fig.~\ref{time_comparison}. The method SCN achieves the best inference efficiency due to its simple architecture, while our AINet introduces more layers and operations, consequently, the inference is slightly slower than the SCN.
The superpixel segmentation of SEAL and SSN is much complex comparing to the SCN and our AINet,
SEAL needs first output the learned deep features and then feed them to a traditional algorithm to conduct superpixel segmentation, and SSN further performs the $K$-means iteration after obtaining the pixel affinity. As a result, SEAL and SSN both cost much more time in inference stage. Although the SCN is faster, its performance is not that satisfactory.  AINet could attain much better performance at the acceptable cost of inference time.   
Comparing to AINet, AINet+ further include one more AI module, consequently, its inference is slightly slower than AINet. Considering the performance gain, the inference cost is acceptable.

\subsection{Application on Object Proposal Generation}
%The superpixels are extremely useful in image annotation 
Image annotation is one of the important application scenarios for superpixels,
since it could identify the semantic boundaries and provide the contours of many semantic regions.  To generate the object proposals, Liu \etal~\cite{DEL} propose a model named DEL, they first estimate the similarities between superpixels and merge them according to a certain threshold, by which the proposed method could flexibly control the grain size of object proposal. In this subsection, we study the quality of object proposals based on the superpixels from different methods to further investigate the superiority of our methods. Specifically,
we first feed the superpixels from five state-of-the-art methods, SEAL, SCN, SSN, AINet and AINet+ to the framework of~\cite{DEL}, according to the estimated similarities, we merge the superpixels according to a threshold to produce object proposals. To evaluate the performance, we use the ASA score to measure how well the produced object proposals cover the ground-truth labels:
\begin{equation}
    ASA(O,G) = \frac{1}{N}\sum_{O_k}\max_{G_k}{\{|O_k\cap G_k|\}},
\end{equation}
where $N$ is the number of generated object proposal $O$, and $G$ is the ground-truth semantic label. 

The performance of all methods is reported in Fig.~\ref{obp}, from which we can observe that the average performance of our AINet and AINet+ is more outstanding than other deep learning-based methods. The models AINet+, AINet, SCN and SSN could be trained in an end-to-end manner, and all perform better than the SEAL, among them, our AINet+ is the most outstanding method.  
Fig.~\ref{del_dis} I shows three results of DEL~\cite{DEL} with the superpixels from four methods,  different thresholds are used to produce varied size proposals: the adjacent superpixels would be merged if their similarity is above the threshold, which means that higher value would produce finer object proposals.  As shown in Fig.~\ref{del_dis} I, our AINet and AINet+ could generate more satisfactory object proposals comparing to the competing methods, which validates the effectiveness of our proposed methods. Fig.~\ref{del_dis} II exhibits the results using the superpixels of our AINet with different thresholds,  varying sizes of generated object proposals could be generated by adjusting the threshold.

\subsection{Application on Stereo Matching}
Besides directly applying for dense annotation task, superpixels could also be integrated into many deep frameworks to provide some guidance, such that some performance gains could be obtained. In this subsection, we conduct experiments to further study the effectiveness of our superpixels on stereo matching task.

Stereo matching seeks to build pixel correspondences between a pair of rectified images. Most recent works promote the performance at the cost of high computation and memory cost~\cite{PSMNet,sm1,sm2}.  Liu~\etal~\cite{SCN} propose to integrate the superpixel segmentation network for memory-saving. Specifically, the input image $I$ is first fed into the superpixel segmentation network to acquire the association map $Q$, then $I$ is down-sampled using Eq.~\ref{step1} and fed forward the stereo matching network for disparity prediction. Finally, the full resolution disparity is obtained by upsampling the disparity using Eq.\ref{step2}, we use the disparity regression  and superpixel losses to joint train the stereo matching and the superpixel networks. More detailed configuration could be found in ~\cite{SCN}.

\setlength{\tabcolsep}{1.5mm}{
\begin{table}[t]
\setlength{\abovecaptionskip}{0pt}%    
\setlength{\belowcaptionskip}{2pt}%
\begin{center}
\begin{tabular}{c|cccc}
\midrule
Method & PSMNet~\cite{PSMNet} & $\text{PSMNet}_{\text{SCN}}$& $\text{PSMNet}_{\text{AINet}}$ &$\text{PSMNet}_{\text{AINet+}}$\\[3pt]
\hline
SceneFlow  &1.04  &0.93 & 0.89&\textbf{0.84}\\[1pt]
HR-VS  & 3.83 &2.77& 2.23 & \textbf{2.02}\\
\midrule
\end{tabular}
\end{center}
%\vspace{-0.2cm}
\caption{The End-point-error (EPE) of four methods on SceneFlow and HR-VS datasets.}
\label{disparity_result}
\vspace{-0.6cm}
\end{table}
}

We conduct experiments on two benchmarks of stereo matching, \ie, SceneFlow~\cite{sceneFlow} and HR-VS~\cite{HRVS}. To make a fair comparison, the experiments on these two datasets follow the same configuration of SCN. We take PSMNet as our baseline method. The End-point-error (EPE) is used for performance evaluation, and  the results are reported in Table~\ref{disparity_result}, where $\text{PSMNet}_{\text{SCN}}$ means the PSMNet equipped with the superpixel network SCN. From Table~\ref{disparity_result}, we can find that the superpixel network could help improve the stereo matching. Comparing to the baseline model PSMNet, introducing the superpixels from SCN, AINet and AINet+ could all boost the performance. For example, when the SCN network is applied to the PSMNet, the EPE on SceneFlow could be improved from 1.04 to 0.93, the AINet and AINet+ make the performance step further, and the PSMNet equipped with our AINet+ achieves the best results. Fig.~\ref{main_visual_disparity} shows the visualization of three methods, from which we could intuitively observe the superiority of our method.

\section{Conclusion}
\label{conclusion}
We have presented an association implantation network for superpixel segmentation task. A novel association implantation module is proposed to provide the consistent pixel-superpixel level context for superpixel segmentation task, we plug the association implantation operation in multiple layers to further boost the performance. By applying our AI module to multi-layers, the model could obtain the hierarchical pixel-superpixel context. Such a design makes the network more effectively infer the association map, bringing more performance gains. To pursue better boundary precision, a boundary-perceiving loss is designed to improve the discrimination of pixels around boundaries in hidden feature level, and a data augmentation named patch jitter is developed to further improve the performance. Experiments on two popular benchmarks show that the proposed method could achieve state-of-the-art performance with good generalizability. What's more, the produced superpixels by our method could also perform well when applied to the object proposal generation and stereo matching tasks. In the future,  we will continue to study the application of superpixel on segmentation task.

\bibliographystyle{ieee_fullname}
\bibliography{main}

% biography section
% 
% If you have an EPS/PDF photo (graphicx package needed) extra braces are
% needed around the contents of the optional argument to biography to prevent
% the LaTeX parser from getting confused when it sees the complicated
% \includegraphics command within an optional argument. (You could create
% your own custom macro containing the \includegraphics command to make things
% simpler here.)
%\begin{IEEEbiography}[{\includegraphics[width=1in,height=1.25in,clip,keepaspectratio]{mshell}}]{Michael Shell}
% or if you just want to reserve a space for a photo:
\begin{IEEEbiography}[{\includegraphics[width=1in,height=1.25in,clip,keepaspectratio]{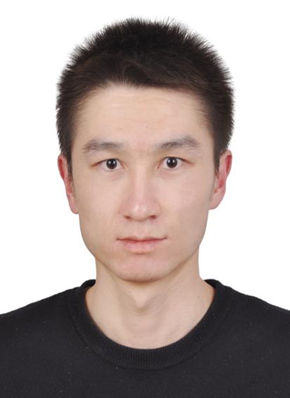}}]{Yaxiong Wang}
received the B.S. degree from Lanzhou University, Lanzhou, China, in 2015. He received his Ph.D degree at School of software Engineering, Xi’an Jiaotong University, Xi’an, China. He is now an assistant professor of Hefei University of Technology, Hefei, China. His current research interests include cross-modal retrieval, image generation, and AIGC detection.
\end{IEEEbiography}

\begin{IEEEbiography}[{\includegraphics[width=1in,height=1.25in,clip,keepaspectratio]{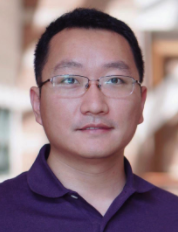}}]{Yunchao Wei}
received the Ph.D. degree from Beijing Jiaotong University, Beijing, China, in 2016.
He is currently an Professor with the Beijing Jiaotong University, Beijing, China. Before joining the Beijing Jiaotong University, he was a lecturer with University of Technology Sydney, Sydney, Australia. He was a Postdoctoral Researcher with the Beckman Institute, University of Illinois at Urbana–Champaign, Urbana, IL, USA, from 2017 to 2019. His current research interests include computer vision and machine learning.
Dr. Wei is an ARC Discovery Early Career Researcher Award Fellow from 2019 to 2021.
\end{IEEEbiography}

\begin{IEEEbiography}[{\includegraphics[width=1in,height=1.25in,clip,keepaspectratio]{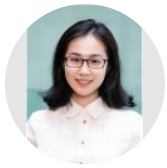}}]{Yujiao Wu}
received the Ph.D degree from University of Technology Sydney, Australia, in 2021. She is now a research scientist in CSRIO, Australia. Her current research interests include cross-modal retrieval, person retrieval, medical data analysis, and object detection.
\end{IEEEbiography}

\begin{IEEEbiography}[{\includegraphics[width=1in,height=1.25in,clip,keepaspectratio]{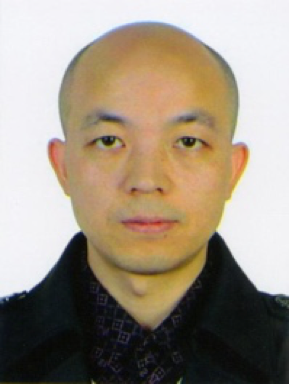}}]{Xueming Qian}
(M’10) received the B.S. and M.S. degrees from the Xi’an University of Technology, Xi’an, China, in 1999 and 2004, respectively, and the Ph.D. degree from the School of Electronics and Information Engineering, Xi’an Jiaotong University, in 2008. He was a Visiting Scholar with Microsoft Research Asia from 2010 to 2011. He was an Assistant Professor with Xi’an Jiaotong University, where he was an Associate Professor from 2011 to 2014, and is currently a Full Professor. He is also the Director of the Smiles Laboratory at Xi’an Jiaotong University. He received the Microsoft Fellowship in 2006. He received outstanding doctoral dissertations of Xi’an Jiaotong University and Shaanxi Province, in 2010 and 2011, respectively. His research interests include social media big data mining and search. His research is supported by the National Natural Science Foundation of China, Microsoft Research, and Ministry of Science and Technology.
\end{IEEEbiography}

\begin{IEEEbiography}[{\includegraphics[width=1in,height=1.25in,clip,keepaspectratio]{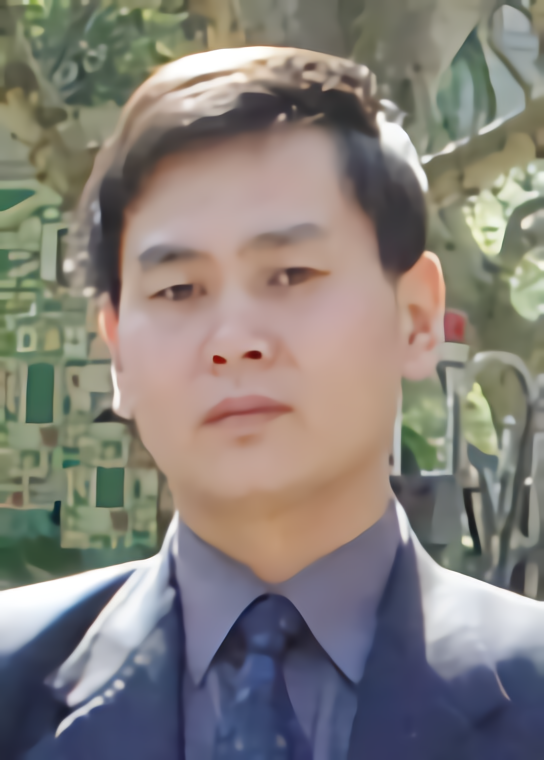}}]{Li Zhu}
 is an associate professor in School of Software, Xi’an Jiao-tong University. He received his M.S. and Ph.D. degrees from Xi’an Jiaotong University in 1995 and 2000, respectively. He received his B.S. degree from Northwestern Polytechnical University in 1989. His main research interests include multimedia processing \& communication, parallel computing and networking.
\end{IEEEbiography}

\begin{IEEEbiography}[{\includegraphics[width=1in,height=1.25in,clip,keepaspectratio]{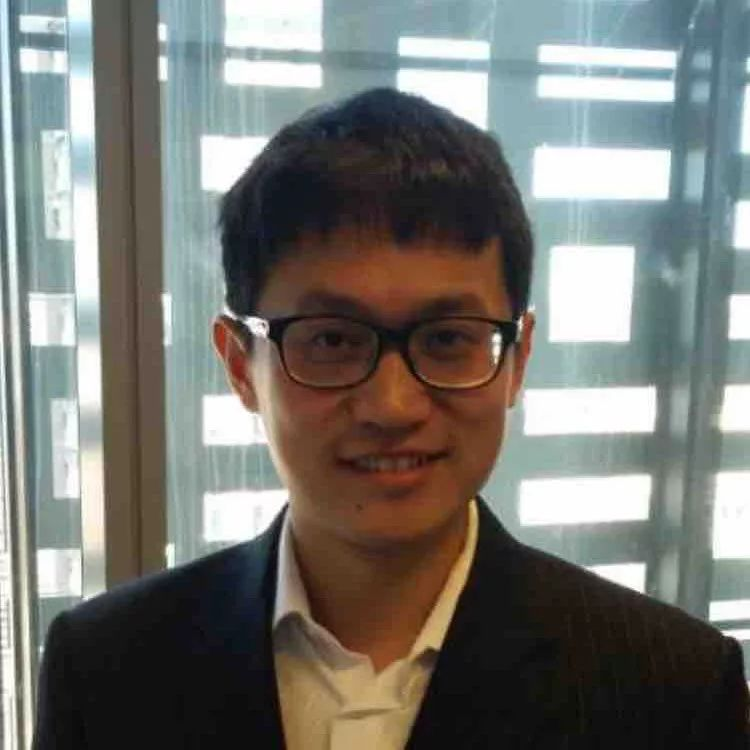}}]{Yi Yang}
 received the Ph.D. degree in computer science from Zhejiang University, Hangzhou, China, in 2010.
He is currently a Professor with Zhejiang University, Hangzhou, China. Before joining the Zhejiang University, he served as a professor with the University of Technology Sydney, Sydney, NSW, Australia. He was a Postdoctoral Researcher with the School of Computer Science, Carnegie Mellon University, Pittsburgh, PA, USA. His current research interests include machine learning and its applications to multimedia content analysis and computer vision, such as multimedia indexing and retrieval, surveillance video analysis, and video semantics understanding.
\end{IEEEbiography}
% You can push biographies down or up by placing
% a \vfill before or after them. The appropriate
% use of \vfill depends on what kind of text is
% on the last page and whether or not the columns
% are being equalized.

%\vfill

% Can be used to pull up biographies so that the bottom of the last one
% is flush with the other column.
%\enlargethispage{-5in}

% that's all folks
\end{document}